\documentclass[10pt,twocolumn,letterpaper]{article}

\usepackage[pagenumbers]{cvpr} % To force page numbers, e.g. for an arXiv version

\usepackage{graphicx}
\usepackage{amsmath}
\usepackage{amssymb}
\usepackage{booktabs}
\usepackage{enumitem}
\usepackage{comment}
\usepackage[accsupp]{axessibility}

\usepackage[pagebackref,breaklinks,colorlinks]{hyperref}

\usepackage[capitalize]{cleveref}
\crefname{section}{Sec.}{Secs.}
\Crefname{section}{Section}{Sections}
\Crefname{table}{Table}{Tables}
\crefname{table}{Tab.}{Tabs.}

\def\cvprPaperID{8138} % *** Enter the CVPR Paper ID here
\def\confName{CVPR}
\def\confYear{2022}

\begin{document}

\title{Long-Short Temporal Contrastive Learning of Video Transformers}

\author{Jue Wang$^{1}$
\hspace{30pt}
Gedas Bertasius$^2$
\hspace{30pt}
Du Tran$^1$
\hspace{30pt}
Lorenzo Torresani$^{1,3}$
\\
$^{1}$Facebook AI Research \hspace{30pt} $^{2}$UNC Chapel Hill \hspace{30pt} $^{3}$Dartmouth
\\
}

\maketitle

\begin{abstract}
Video transformers have recently emerged as a competitive alternative to 3D CNNs for video understanding. However, due to their large number of parameters and reduced inductive biases, these models require supervised pretraining on large-scale image datasets to achieve top performance. In this paper, we empirically demonstrate that self-supervised pretraining of video transformers on video-only datasets can lead to action recognition results that are on par or better than those obtained with supervised pretraining on large-scale image datasets, even massive ones such as ImageNet-21K. Since transformer-based models are effective at capturing dependencies over extended temporal spans, we propose a simple learning procedure that forces the model to match a long-term view to a short-term view of the same video. Our approach, named Long-Short Temporal Contrastive Learning (LSTCL), enables video transformers to learn an effective clip-level representation by predicting temporal context captured from a longer temporal extent. To demonstrate the generality of our findings, we implement and validate our approach under three different self-supervised contrastive learning frameworks (MoCo v3, BYOL, SimSiam) using two distinct video-transformer architectures, including an improved variant of the Swin Transformer augmented with space-time attention. We conduct a thorough ablation study and show that LSTCL achieves competitive performance on multiple video benchmarks and represents a convincing alternative to supervised image-based pretraining. 
    
\end{abstract}
\section{Introduction}
\label{sec:intro}
Since the introduction of AlexNet~\cite{krizhevsky2012imagenet}, deep convolutional neural networks (CNNs) have emerged as the prominent model in numerous computer vision tasks~\cite{he2016deep,goodfellow2016deep,simonyan2014two,goodfellow2014generative,wang2019gods,wang2018learning}. More recently, the Transformer model~\cite{vaswani2017attention} has received much attention due to its impressive performance in the field of natural language processing (NLP)~\cite{devlin2018bert}. While CNNs rely on the local operation of convolution, the building block of transformers is self-attention~\cite{vaswani2017attention} which is particularly effective at modelling long-range dependencies. In the image domain, the Vision Transformer (ViT)~\cite{dosovitskiy2020image} was proposed as a convolution-free architecture which uses self-attention between non-overlapping patches in all layers of the model. ViT was shown to be competitive with state-of-the-art CNNs on the task of image categorization. In the last few months, several adaptations of ViT to video have been proposed~\cite{bertasius2021space,neimark2021video,arnab2021vivit}. In order to capture salient temporal information from the video, these works typically extend the self-attention mechanism to operate along the time axis in addition to within each frame. Since video transformers have a larger numbers of parameters and fewer inductive biases compared to CNNs, they typically require large-scale pretraining on supervised image datasets, such as ImageNet-21K~\cite{ridnik2021imagenet} or JFT~\cite{arnab2021vivit}, in order to achieve top performance.

Self-supervised learning has been shown to be an effective solution to eliminate the need for large-scale supervised pretraining of transformers both in NLP~\cite{devlin2018bert} as well as in image-analysis~\cite{touvron2020training,caron2021emerging}. In this work, we show that, even in the video domain, self-supervised learning provides an effective way for pretraining video transformers. Specifically, we introduce Long-Short Temporal Contrastive Learning (LSTCL), a contrastive formulation that maximizes representation similarity between a long video clip (say, 8 seconds long) and a much shorter clip (say, 2 seconds long) where both clips are sampled from the same video. We argue that by training the short-clip representation to match the long-clip representation, the model is forced to extrapolate from a short extent the contextual information exhibited in the longer temporal span. As the long clip includes temporal segments not included in the short clip, this self-supervised strategy trains the model to anticipate the future and to predict the past from a small temporal window in order to match the representation extracted from the long clip. We believe that this is a good pretext for video representation learning, as it can be accomplished only by a successful understanding and recognition of the structure and correlation of atomic actions in a long video. Furthermore, such framework is particularly suitable for video transformers as they have been recently shown to effectively capture long-term temporal cues~\cite{bertasius2021space}. In this work we demonstrate that these long-term temporal cues can be effectively encoded into a short-range clip-level representation leading to a substantial improvement in video classification performance.

To demonstrate the generality of our findings, we experiment with two different video transformer architectures whose code is publicly available. The first is TimeSformer~\cite{bertasius2021space}, which reduces the computational cost of self-attention over the 3D video volume by means of a space-time factorization. The second architecture is the Swin transformer~\cite{liu2021swin}, which we further extend into a 3D version, dubbed Space-Time Swin transformer, that computes hierarchical spatiotemporal self-attention by using 3D shifting windows. We show that our unsupervised LSTCL pretraining scheme allows both of these video transformers to outperform their respective counterparts pretrained with full supervision on the large-scale ImageNet-21K dataset.%\GB{End.}

In summary, the contributions of this paper can be summarized as follows:
\begin{itemize}[leftmargin=*,topsep=0pt,noitemsep]
%[leftmargin=*][
\item We introduce Long-Short Temporal Contrastive Learning (LSTCL), which enables encoding temporal context from the longer video into a short-range clip representation.
\item We demonstrate that for recent video transformer models, our proposed LSTCL pretraining provides an effective alternative to large-scale supervised pretraining on images.
\item We propose a Space-Time Swin transformer for spatiotemporal feature learning, and show that it achieves strong results on multiple action recognition benchmarks. 
%\item Our work achieve promising performance in multiple benchmarks, such as Kinetics-400, Kinetics-600 and Something-Something-V2, demonstrating that video transformer models with self-supervised pretraining is able to outperform the same model with ImageNet pretraining.
\end{itemize}
\section{Related work}
\noindent\textbf{Self-supervised Learning in Images.} Early attempts at self-supervised visual representation learning used a variety of pretext tasks, such as image rotation prediction~\cite{komodakis2018unsupervised}, auto-encoder learning~\cite{vincent2008extracting, pathak2016context,song2018self}, or solving jigsaw puzzles~\cite{noroozi2016unsupervised}. In comparison, recent approaches in self-supervised learning leverage contrastive learning~\cite{he2020momentum, chen2020simple, chen2020improved,chen2021empirical,caron2021emerging,recasens2021broaden,Han20}.  The idea is to generate two views of the same image through data augmentation and then minimize the distance of their representations while, optionally, maximizing the distance to other images~\cite{chen2020simple, he2020momentum}. One disadvantage of contrastive learning is that it requires a large number of negative examples, which implies a large batch size~\cite{chen2020simple} or the use of a memory bank~\cite{he2020momentum}. To tackle the high computational cost of such contrastive approaches, several recent methods proposed to eliminate the reliance on negative samples~\cite{chen2020exploring,chen2021empirical,grill2020bootstrap,caron2020unsupervised}. %In this paper, we experiment with three popular self-supervised learning frameworks proposed for the image-domain: BYOL~\cite{grill2020bootstrap}, MoCo v3~\cite{chen2021empirical} and SimSiam~\cite{chen2020exploring}. 

%However, the key of all frameworks mentioned above is the same, that is to do the alignment between features from similar samples while maintaining the uniformity of feature distribution on the output unit hyper-sphere~\cite{wang2020understanding}. In this paper, we adopt three of the self-supervised learning frameworks that do not require negative instances discrimination, they are BYOL~\cite{grill2020bootstrap}, MoCo v3~\cite{chen2021empirical} and SimSiam~\cite{chen2020exploring}. 

\noindent\textbf{Self-supervised Learning in Videos.} Several methods for self-supervised video representation learning focused on predictive spatiotemporal ordering tasks~\cite{goroshin2015unsupervised,DBLP:journals/corr/IsolaZKA15,Agrawal_2015_ICCV, pmlr-v37-srivastava15, 7410677, Misra-2016-5596, xu2019self, tao2020self, liu2022tcgl, yao2021seco, hu2021contrast}. Other approaches have leveraged temporal cues such as tempo and speed to define self-supervised pretext tasks~\cite{benaim2020speednet, wang2020self}. Just like in the image domain, more recent approaches~\cite{Feichtenhofer_large, han2019video, DBLP:journals/corr/abs-2008-03800} adopt contrastive learning objectives. Our method also falls into the category of contrastive approaches. In comparison to prior contrastive video methods, we propose a contrastive formulation where a positive pair is generated from a short clip and a long clip, both of which are sampled from the same video. This pushes our model to learn a short clip-level representation that captures global video-level context. 
%Furthermore, we note that most prior self-supervised video learning approaches~\cite{piergiovanni2020evolving,Han20,benaim2020speednet,han2019video} pretrain models on large-scale datasets (e.g., Kinetics or YouTube-8M), and then they evaluate them on dramatically smaller-scale benchmarks, such as HMDB51, UCF101 or a small subset of Kinetics. The problem we address is different: it entails video classification {\em without} the use of additional data. It is more challenging than classification in small benchmarks by means of transfer learning from large datasets.

The approach that is most closely related to our own is the BraVe system~\cite{recasens2021broaden}. BraVe shares the same underlying idea of training a model to match a long (broad) view to a short (narrow) view of the same video. However, our work differs in several aspects. First of all, our main focus is to leverage self-supervised learning as a means to train video transformers without labeled image data, while BraVe is applied to 3D CNNs. Video transformers are emerging as a competitive alternative to 3D CNNs. However, as discussed earlier, they suffer from the limitation of requiring image-based supervised pretraining. Thus, we believe that this is an important and timely problem to address. Additionally, we note that our LSTCL is a lot simpler than BraVe: while our model uses shared parameters, a single projection network, and a single prediction network, BraVe requires separate backbones, separate projection networks, and separate prediction networks for the two views in order to achieve the best performance; furthermore, while LSTCL can be applied with any traditional contrastive loss (as demonstrated by our experiments with MoCo v3, BYOL, and SimSiam), BraVe uses a combination of two specific regression objectives (broad-to-narrow and narrow-to-broad) and employs distinct augmentation strategies for the two views. Despite the bare-bone simplicity of our learning formulation, we demonstrate that it delivers impressive results, elevating the accuracy of video transformers to the state-of-the-art on challenging action classification benchmarks {\em without} the need of any supervised image-level pretraining.

\noindent\textbf{Transformers in Vision.} Transformer-based models~\cite{vaswani2017attention,devlin2018bert} currently define the state-of-the-art for the majority of natural language processing (NLP) tasks. Similarly, there have also been several attempts to adopt transformer-based architectures for vision problems. Initially, these attempts focused on architectures mixing convolution with self-attention~\cite{wang2018non,wang2020end,carion2020end,huang2019ccnet,zhang2020dynamic}. The recent introduction of Vision Transformer (ViT)~\cite{dosovitskiy2020image} has demonstrated that it is possible to achieve competitive image classification results with a convolution-free architecture. To increase the data-efficiency aspect of the original ViT, Touvron et al.~\cite{touvron2020training} proposed a training recipe based on distillation. Lastly, the recently introduced Swin transformer~\cite{liu2021swin} significantly reduces the number of parameters and the cost of ViT by employing local rather than global self-attention.

ViT models were also adapted to the video domain by introducing different forms of spatiotemporal self-attention~\cite{bertasius2021space, arnab2021vivit,akbari2021vatt,patrick2021keeping}. However, due to their large number of parameters, these models typically require large amounts of training data, which typically comes in the form of a large scale labeled dataset (such as ImageNet or JFT). To address this issue, Fan et al.~\cite{fan2021multiscale} introduced a multi-scale vision transformer (MViT), which uses a much smaller number of parameters and can be successfully trained from scratch. Instead of reducing the model capacity, as done in MViT, we show that it is possible to train large-capacity video transformer models without any external data by means of our proposed LSTCL self-supervised learning framework.
\section{Video Transformers}
% \JW{removed 'Background' in the section title, as ST Swin is a new extension model proposed here.}\\
%\subsection{Video Transformer Model}
Several recent attempts have been made to extend ViT to the video domain~\cite{bertasius2021space, arnab2021vivit, fan2021multiscale, akbari2021vatt,patrick2021keeping}. Most video transformers share common principles, which we review below. We then discuss specific designs differentiating the video transformers considered in our experiments. %, which can be roughly defined as: 1) Video decomposition, 2) Linear and position embedding, 3) Multi-head attention encoding, 4) Classification.  

\subsection{Overview}

\noindent\textbf{Linear and positional embeddings.} Each patch $\boldsymbol{p}_{(i,t)}$ is linearly embedded into a feature vector $\smash{\boldsymbol{z}^{0}_{(i,t)} \in \mathbb{R}^D}$ obtained by means of a learnable matrix $\smash{W \in\mathbb{R}^{D \times (P^2 \cdot C)}}$ and a learnable vector $\smash{\boldsymbol{e}_{(i,t)}\in\mathbb{R}^D}$ representing the spatial-temporal positional embedding: $\smash{\boldsymbol{z}^{0}_{(i,t)} = W \boldsymbol{p}_{(i,t)} + \boldsymbol{e}_{(i,t)}}$. %Similar to BERT~\cite{devlin2018bert}, a classification token $v_{(0,0)}$ will also be added at the beginning of the input sequence. for outputting final classification probabilities in the last layer of the transformer.

\noindent\textbf{Multi-headed attention.} The multi-headed self-attention (MHA) is the key component of the transformer. It implements the query-key-value computation for each patch, and it is interleaved with layer normalization~\cite{ba2016layer} (LN) and a multilayer perceptron (MLP) within each block $\ell$. Thus, the intermediate representation $\boldsymbol{z}^\ell$ for a patch in block $\ell$ is obtained from its features in the previous block, as: %More details can be found in ~\cite{devlin2018bert,vaswani2017attention}. Most video transformer methods follow similar architecture that can be written as:

\begin{align}
    &\tilde{\boldsymbol{z}}^\ell=\text{MHA}(\text{LN}(\boldsymbol{z}^{\ell-1}))+\boldsymbol{z}^{\ell-1}\\
    &\boldsymbol{z}^\ell = \text{MLP}(\text{LN}(\tilde{\boldsymbol{z}}^\ell))+\tilde{\boldsymbol{z}}^\ell~.
    \label{mha}
\end{align}
%Where MHA, LN and MLP represent multi-head attention, layer normalization and multilayer perceptron. $\boldsymbol{z}^L$ is the intermediate feature from $L^{th}$ layer.

\noindent\textbf{Classification.} As in BERT~\cite{devlin2018bert}, a classification token $\boldsymbol{p}_{(0,0)}$ is added at the beginning of the input sequence. In the last layer of the network, a linear layer with softmax activation function is attached to the classification token in order to output the final classification probabilities.

%In this paper, we investigate our study on two different video transformers. They are TimeSformer~\cite{bertasius2021space} and  Space-Time Swin transformer. 

%\textbf{TimeSformer~\cite{bertasius2021space}.}  

\subsection{TimeSformer}

\begin{figure}[]
\begin{center}
   \includegraphics[width=1.0\linewidth]{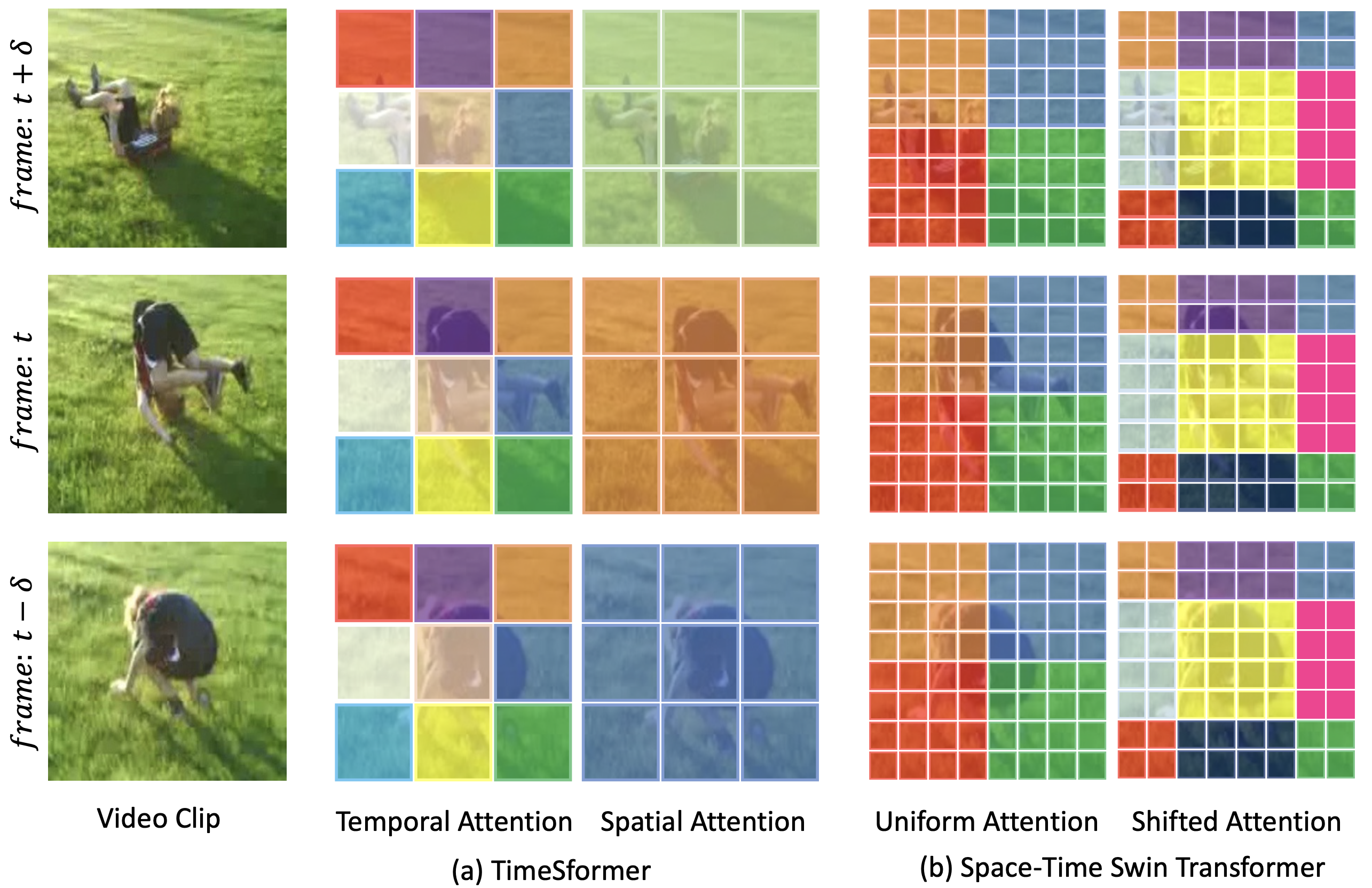}
\end{center}
   \caption{An illustration of the self-attention mechanisms in TimeSformer~\cite{bertasius2021space} and Space-Time (ST) Swin Transformer. Each column in the figure depicts a different self-attention block. Patches that have the same color are compared during the self-attention computation.}
%   \GB{1) the captions above each frame would look better on the left side of each frame (positioned vertically from bottom to top. 2) Furthermore, it would look better to have more vertical space between captions depicting the attention mechanism, and the method. The methods could be listed as a) TimeSformer, b) Space-Time Swin Transformer. 3) More horizontal space betweeen video clip column, TimeSformer column, and Space-Time Swin Transformer column would make the figure clearer. 4) Lastly, please fix the typos.}\JW{done} }
\label{fig:atten_map}
\end{figure}

TimeSformer~\cite{bertasius2021space} extends ViT~\cite{dosovitskiy2020image} to the video domain. It uses two independent multi-head attention blocks for spatial and temporal self-attention. As shown in Figure~\ref{fig:atten_map}, the spatial self-attention compares the query patch only to image patches appearing in the same frame. Conversely, the temporal self-attention compares the query patch to the image patches in the same spatial location but from the other frames. The decomposition over space and time dramatically reduces the cost of self-attention compared to a dense comparison over all pairs of patches of the video. Thus, the feature representation is computed as:

%In TimeSformer, the Equation~\ref{mha} can be re-written as:

%Compared to the image-based vision transformer, TimeSformer extensively investigate how to cooperate the temporal self-attention in the vision transformer. Through the ablation study of several  spatial-temporal self-attention designs, we pick the best-performed one in this paper, namely TimeSformer with divided space-time attention. In which, it proposes two independent multi-head attention block for doing the self-attention spatially and temporally. As is shown in the Figure~\ref{fig:atten_map}, it learns spatial self-attention by using image patches frame-wisely while learns temporal self-attention by using image patches position-wisely in the video clip. In TimeSformer, the Equation~\ref{mha} can be re-written as:
\begin{align}
    &\boldsymbol{z}^\ell_t=\text{MHA}_{Time}(LN(\boldsymbol{z}^{\ell-1}))+\boldsymbol{z}^{\ell-1} \\\nonumber
    &\boldsymbol{z}^\ell_s=\text{MHA}_{Space}(LN(\boldsymbol{z}^\ell_t))+\boldsymbol{z}^\ell_t \\ \nonumber
    &\boldsymbol{z}^\ell = \text{MLP}(\text{LN}(\boldsymbol{z}^\ell_s))+\boldsymbol{z}^\ell_s \nonumber
\end{align}

\subsection{Space-Time Swin Transformer}
\label{ST Swin}
% \JW{should we expand bit m ore here? given ST Swin is one of our contributions.}
% \textbf{Space-Time Swin Transformer.}
% \GB{Sometimes we refer to this as a 3D Swin transformer. I think it would be good to pick a name and use it consistently} 

Compared to ViT, the Swin Transformer~\cite{liu2021swin} applies self-attention locally. The features are learned hierarchically by aggregating information from local neighborhoods of patches in each layer. Here we adapt the original Swin transformer, which was introduced for still-images, to video. We name this new variant Space-Time Swin Transformer (ST Swin). Instead of considering 2D neighborhoods of image patches for self-attention computation, ST Swin uses local 3D space-time volumes. Specifically, as proposed in the original paper~\cite{liu2021swin}, ST Swin uses two distinct self-attention mechanisms: uniform partition and shifted partition. In our case, both of these self-attention schemes are adapted to video by considering the temporal dimension in the local patch neighborhoods. As shown in Figure~\ref{fig:atten_map}, the uniform partition splits the entire clip into 4 non-overlapping 3D sections, with each section sharing the same partition index. Spatiotemporal self-attention is then computed between image patches that have the same partition index. Similarly, shifted partition generates multiple non-overlapping 3D sections at different scales, and spatiotemporal patches within each section are compared for self-attention computation. The uniform partition and the shifted partition are stacked to form two successive attention blocks, which implement cross-window connections further increasing the model capacity. Thus, the complete transformation carried out in each layer $\ell$ of the ST Swin transformer can be summarized as follows::
\begin{align}
    \boldsymbol{z}^\ell_u&=\text{MHA}_{Uniform}(LN(\boldsymbol{z}^{\ell-1}))+\boldsymbol{z}^{\ell-1} \\\nonumber
    \boldsymbol{z}^\ell &= \text{MLP}(\text{LN}(\boldsymbol{z}^\ell_u))+\boldsymbol{z}^\ell_u \\ \nonumber
    \boldsymbol{z}^{\ell+1}_s&=\text{MHA}_{Shift}(\text{LN}(\boldsymbol{z}^\ell))+\boldsymbol{z}^\ell \\ \nonumber
    \boldsymbol{z}^{\ell+1} &= \text{MLP}(\text{LN}(\boldsymbol{z}^{\ell+1}_s))+\boldsymbol{z}^{\ell+1}_s \nonumber
\end{align}
We adopt the 3D relative positional embedding and the patch merging strategy used in Swin~\cite{liu2021swin}. However, we only merge image patches along the spatial axis while maintaining fixed temporal resolution through the layers. 

%\LT{Can we explain in further details what the uniform and shifted partitions are, and how we extend the neighborhoods along the temporal axis?}\JW{expand the context of ST Swin a bit, maybe we could remove eq1-3 as they are not very important}

%Given spatiotemporal image patches, both TimeSformer~\cite{bertasius2021space} and ViViT~\cite{arnab2021vivit} study different designs of spatial and temporal attention blocks. From their experimental results as well as to save computational cost, we choose to learn spatiotemporal attention jointly in this paper. In addition, we continue two window partition strategies mentioned in~\cite{liu2021swin}: uniform partition and shifted partition, and adopt it in our video based set-up. The detail of ST Swin attention mechanism is demonstrated in the Figure~\ref{fig:atten_map}. 

% \DT{why do we use $z^{L+1}$ instead of $z^{\ell+1}$ in the last two equations of Eq. (3)? Do I miss something?}

\section{Long-Short Temporal Contrastive Learning}
\label{LSTCLsec}

\noindent\textbf{Overview.} Video transformers have been shown to be particularly effective at long-range temporal modeling~\cite{bertasius2021space}. Our aim is to design a contrastive learning framework that exploits this characteristic. Our proposed Long-Short Temporal Contrastive Learning (LSTCL) framework takes as input a pair of clips sampled from the same video--a long clip and a short clip. The procedure trains the video transformer to match the representation of the short clip to the representation of the long clip. This forces the model to predict the future and the past from a small temporal window, which is beneficial for capturing the general structure of the video. Below we describe specific details related to our LSTCL.

Given a batch $B$ of {\em unlabeled} training videos, we randomly sample a short clip and a long clip from each of them. While both the long and the short clip include a total of $T$ frames, we use largely different sampling temporal strides $\tau_{S}$ and $\tau_{L}$ with $\tau_S < \tau_L$ in order  for the long clip to cover a much longer temporal extent than the short clip. The sets of short and long clips in the batch $B$ are denoted as $X_S=\{x_S^1,x_S^2,...x_S^B\}$ and $X_L=\{x_L^1,x_L^2,...x_L^B\}$, respectively, where $x_S^i$ and $x_L^i$ represent the short clip and the long clip sampled from the $i$-th example in the batch. The set of short clips is processed by  an encoder $f_q$ to yield a set of ``query'' examples $Q = \{q^1,q^2,...q^B\}$ where $q^i = f_q(x_S^i)\in\mathbb{R}^{D}$. The set of long clips is processed by a separate encoder $f_k$ to produce ``key'' examples $K = \{k^1,k^2,...k^B\}$. We optimize the encoders to yield similar query-key representations for pairs consisting of a long clip and a short clip taken from the same video, and dissimilar representations for cases where the long clip and the short clip are sampled from different videos. This is achieved by adopting an InfoNCE~\cite{oord2018representation} loss on the sets $Q$ and $K$:
\begin{equation}
    \mathcal{L}_{NCE} = \sum_{i} -log\frac{exp({q^i}^\top k^i/\rho)}{exp({q^i}^\top k^i/\rho)+\sum_{j \neq i}exp({q^i}^\top k^j/\rho)}
\end{equation}
%Where $k^+$ and $k^-$ are the output of $f_k$. $k^+$ and $q$ are the representation of two clips that come from the same video, named as positive pairs. $k^-$ is the representation of clips from different videos, known as the negative instance of $q$. 
where $\rho$ is a temperature hyperparameter that controls the sharpness of the output distribution. As commonly done~\cite{chen2021empirical, chen2020exploring, grill2020bootstrap,caron2020unsupervised}, we symmetrize the loss function. In our case this is achieved by adding to the loss term above a dual term obtained by reversing the role of the long and the short clips, i.e., by computing queries from long clips $q^i = f_q(x_L^i)$ and keys from short clips $k^i = f_k(x_S^i)$.  The encoder $f_q$ consists of a video transformer backbone, a MLP projection head and an additional prediction MLP head. The purpose of the prediction layer is to transform the representation of the query clip to match the key. The encoder $f_k$ consists of a video transformer backbone and a MLP projection head. Our experiments present results obtained with different contrastive learning optimizations to update the parameters of $f_q$ and $f_k$. In the case of our default optimization based on MoCo v3~\cite{chen2021empirical}, the parameters of $f_q$ are updated by minimizing $\mathcal{L}_{NCE}$ via backpropagation, while the parameters of $f_k$ are updated as a moving average of the parameters of $f_q$. We refer the reader to our supplementary materials for details of the optimizations based on the other contrastive learning frameworks considered in our experiments---BYOL and SimSiam.

%We adopt the InfoNCE~\cite{oord2018representation} as our objective to minimize. For any $q\in Q$, we have the following loss function in our LSTCL:
%\begin{equation}
%    \mathcal{L}_{q} = -log\frac{exp(qk^+/\rho)}{exp(qk^+/\rho)+\sum\limits_{k^-}exp(qk^-/\rho)}
%\end{equation}
%Where $k^+$ and $k^-$ are the output of $f_k$. $k^+$ and $q$ are the representation of two clips that come from the same video, named as positive pairs. $k^-$ is the representation of clips from different videos, known as the negative instance of $q$. $\rho$ is a temperature parameter that controls the sharpness of the output distribution. 
\begin{figure*}[]
\centering
\begin{subfigure}{.24\textwidth}
  \includegraphics[width=0.8\linewidth]{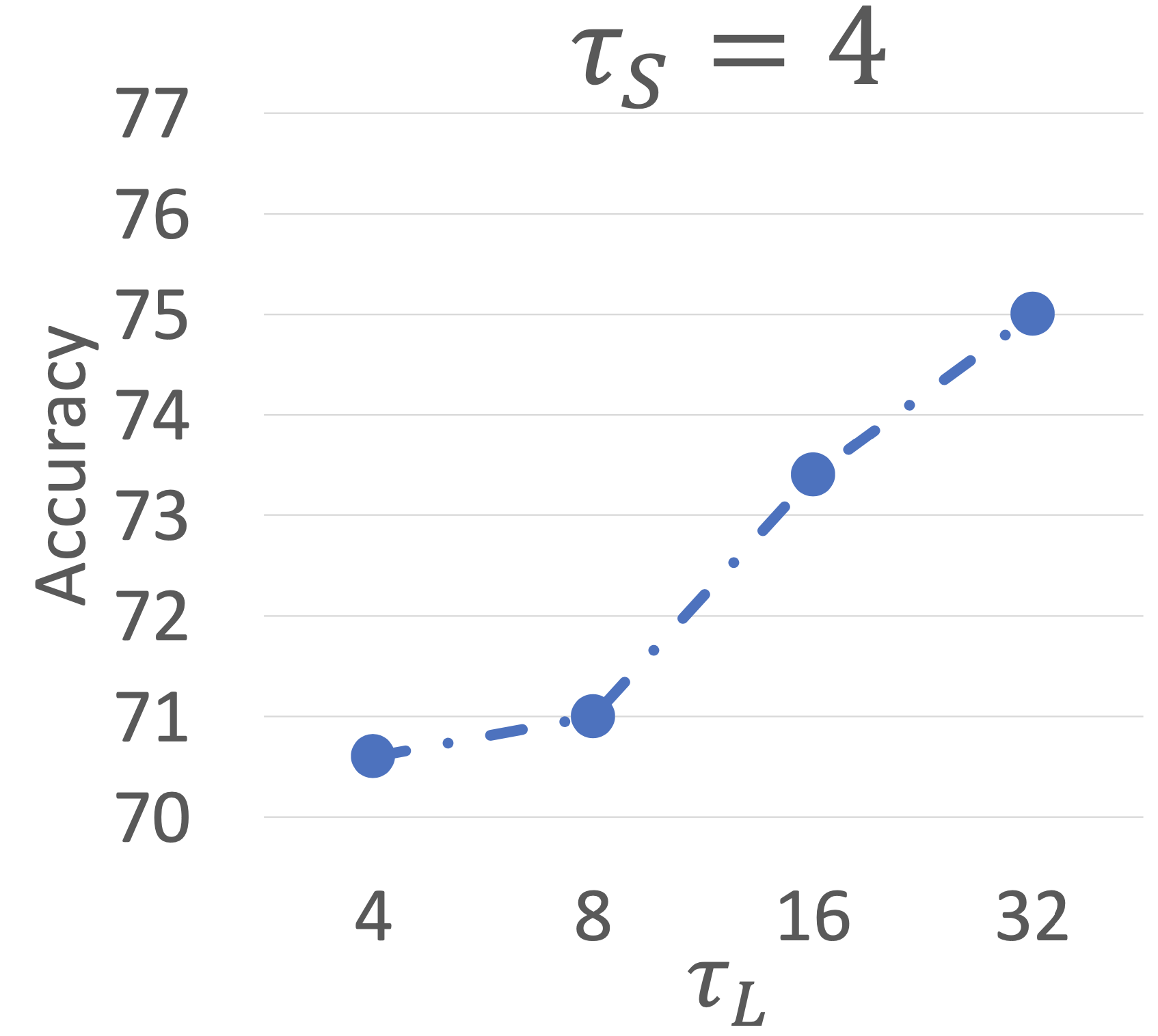}
\end{subfigure}
\begin{subfigure}{.24\textwidth}
  \includegraphics[width=0.8\linewidth]{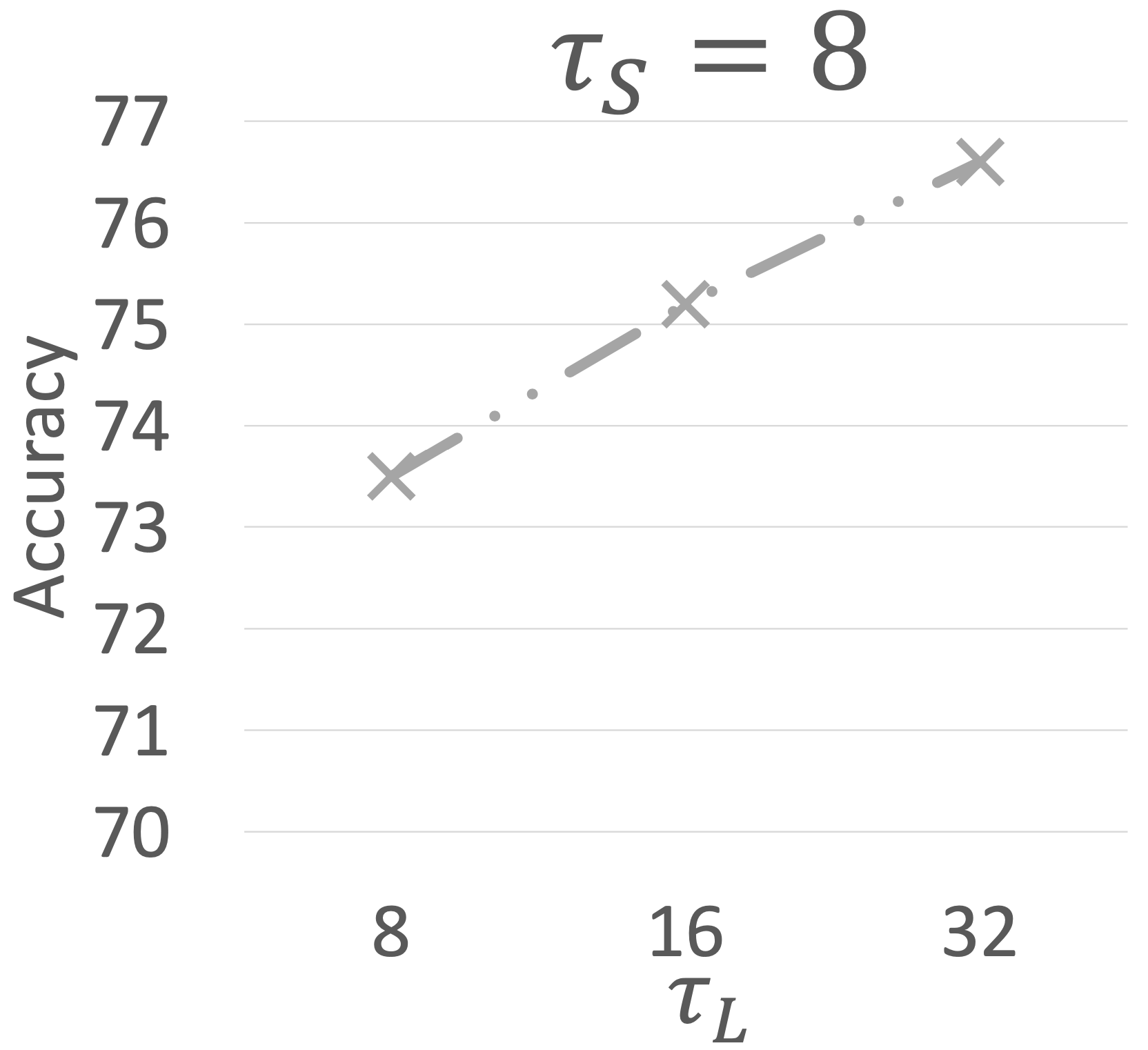}
\end{subfigure}
\begin{subfigure}{.24\textwidth}
  \includegraphics[width=0.8\linewidth]{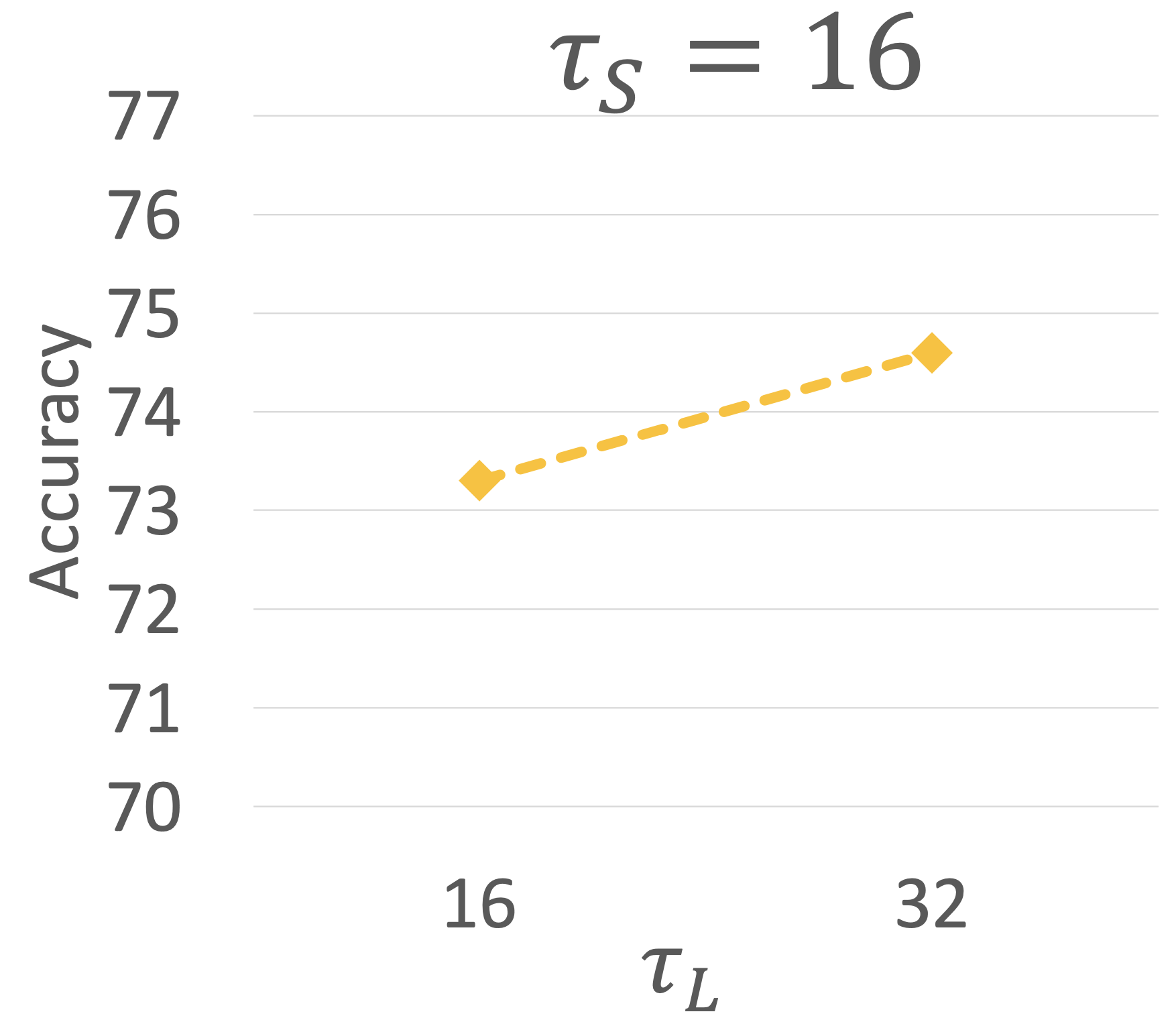}
\end{subfigure}
\begin{subfigure}{.24\textwidth}
  \includegraphics[width=0.8\linewidth]{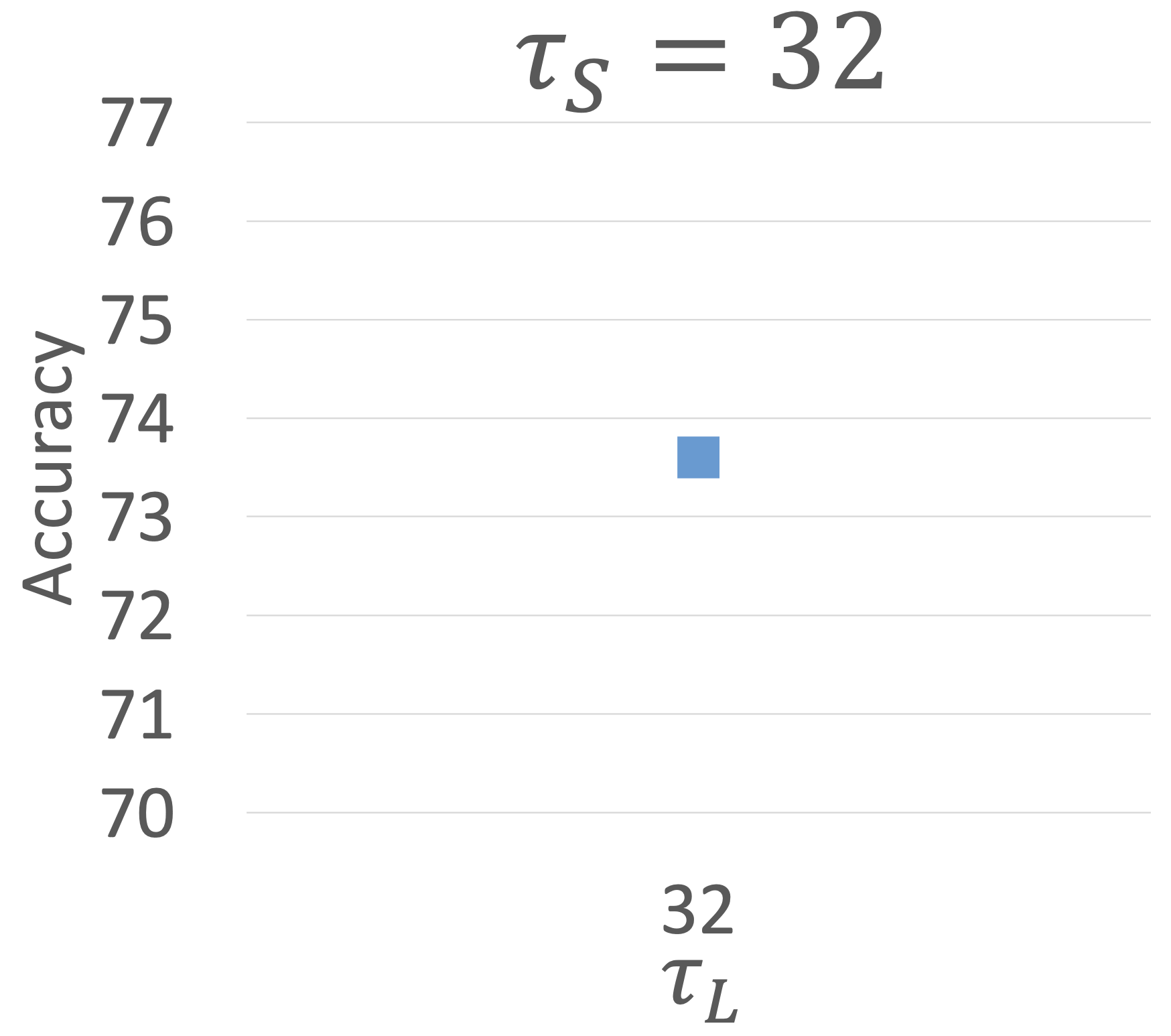}
\end{subfigure}
  \caption{We study how the temporal extents of the short and the long view (controlled by $\tau_S$ and $\tau_L$) in LSTCL affect the video-level accuracy on Kinetics-400. We can see that, for each choice of $\tau_S$, accuracy monotonically increases as the long stride $\tau_L$ is made larger. The best result is obtained for $\tau_S=8$ and $\tau_L=32$, corresponding to a long view that is 4 times longer than the short view.}
\label{fig:sampling rate}
\end{figure*}

\noindent\textbf{Clip Sampling Strategy.} Since we want our model to be able to extrapolate the context observed in the entire video from the brief extent of the short clip, we propose to sample the long and the short clip {\em at random} and {\em independently} from each video. By doing so, the learning cannot leverage any synchrony between the two clips and because the temporal offset will be random for every pair of long-short samples, the optimization will force the short clip representation to encode as much as possible of the context exhibited over the entire video. To demonstrate the value of random independent sampling, in our ablation study we contrast this strategy (named ``Random Independent'') to two alternative schemes. The first, named ``Random Included,'' consists in sampling the short clip at random but so as to fall completely within the temporal extent spanned by the long clip (which is sampled first at random). The second, named ``Random Disjoint,'' samples the two clips at random but it enforces the constraint that they cannot overlap at all, i.e., they are completely disjoint. We refer the reader to our experiments which validate our hypothesis that random independent sampling is indeed the superior strategy for long-short temporal contrastive learning of video transformers. 

%\subsection{Implementation Details}
%\label{implement}

\noindent\textbf{Implementation Details.} We implement LSTCL under three different and popular contrastive learning frameworks: BYOL~\cite{grill2020bootstrap}, MoCo v3~\cite{chen2021empirical}, and SimSiam~\cite{chen2020exploring}. For training we adopt the video data augmentations described in~\cite{Feichtenhofer_large} using clips of size $224 \times 224 \times 8$ sampled from the video. We experiment with two video transformer architectures: TimeSformer with Divided Space-Time attention~\cite{bertasius2021space} and our adaptation of the Swin-B model~\cite{liu2021swin} to video  (Space-Time Swin).  We use the AdamW~\cite{loshchilov2017decoupled} optimizer, which is commonly used for training vision transformer models~\cite{bertasius2021space,caron2021emerging,chen2021empirical,arnab2021vivit,akbari2021vatt,dosovitskiy2020image,touvron2020training}. In our default set-up, we train LSTCL for $200$ epochs on the 240K videos of Kinetics-400~\cite{kay2017kinetics} using linear warm-up~\cite{goyal2017accurate} during the first $40$ epochs. We apply a cosine decay schedule~\cite{loshchilov2016sgdr} after the warm-up and the learning rate is set to $lr \times BatchSize/256$. We adopt the base learning rate and weight decay from~\cite{chen2021empirical}. Our experiments are run on 64 V100 GPUs with a distributed training set-up in Pytorch~\cite{paszke2019pytorch}. The training of 200 epochs takes about three days. 

\section{Experiments}
\label{sec:exp}

We evaluate our proposed LSTCL on several action recognition benchmarks: Kinetics-400~\cite{kay2017kinetics},  Kinetics-600~\cite{carreira2017quo}, Something-Something-V2~\cite{goyal2017something} (SSv2), HMDB~\cite{Kuehne11}, and UCF101~\cite{UCF101}. Our experimental setup is as follows. First, we perform self-supervised LSTCL pretraining on Kinetics-400 with clips of $T=8$ frames but using {\em distinct} temporal sampling strides for the short view and the long view, so that the two views effectively span temporal extents of different lengths in seconds. Afterwards, we finetune the LSTCL-pretrained model for 200 epochs in a fully supervised fashion on each of these three datasets. During inference, we sample uniformly 5 clip with center cropping from each video and average the sample-level predictions to perform video-level classification. In the following ablation studies, unless otherwise noted, we adopt TimeSformer as the backbone in our LSTCL with an input clip of size $8\times224\times224$.

%For finetuning, we adopt the data argumentation and training recipe from~\cite{touvron2020training}. %Unless otherwise noted, all experimental results in the ablation study is the fine-tuning performance in the Kinetics-400 dataset. 

%In addition to Kinetics-400, we present results on 2 other classification datasets: Something-Something-V2~\cite{goyal2017something} (SSv2) and Kinetics-600 datasets~\cite{carreira2017quo}.  Something-Something-V2 dataset includes 108K video for action recognition of 174 categories. While spatial and appearance information are the primary cues needed to achieve strong accuracy on Kinetics-400~\cite{sevilla2021only}, the Something-Something-V2 benchmark defines categories requiring thorough temporal analysis to be properly recognized.  The Kinetics-600 dataset is an expanded version of Kinetics-400. It includes 496K videos over 600 action classes.

%In additional to the Kinetics-400 dataset, we also fine-tune our model from LSTCL in Something-Something-V2~\cite{goyal2017something} (SSv2) and Kinetics-600 datasets~\cite{carreira2017quo}. Specifically, 

\noindent\subsection{Ablation Studies}
\label{ablate}
% \textbf{Learning Rate:} As is mentioned in the ~\cite{chen2021empirical}, vision transformer models are normally sensitive to the learning rate. In the Table~\ref{LR}, we explore a few options of learning rate in the self-supervised learning with TimeSformer. As suggested from the result, the training will become fragile if the learning rate is too large. From the ~\cite{Feichtenhofer_large}, there is no big gap between the learning rate used in different self-supervised frameworks. Also video transformers in this paper share the similar architecture. As a result, we won't further explore the choice of learning rate when using other self-supervised frameworks with different video transformers. Learning rate of $0.001$ will be our default setting in the following experiment.
% \begin{table}[bp]
% \centering
% \caption{Linear evaluation of different base learning rate in the self-supervised video transformer learning. We adopt MoCoV3 with TimeSformer and only one time scale, 8x8, is used. }
% \begin{tabular}{|c|c|c|c|}
% \hline
% Learning Rate & 5e-4  & 1e-3  & 5e-3 \\ \hline
% Linear Evaluation             & 58.1      & 58.5  &  Not converging        \\ \hline
% \end{tabular}
% \label{LR}
% \end{table}

\noindent\textbf{Importance of the Temporal Extent.}
We first ablate the choice of $\tau_S$ and $\tau_L$ for self-supervised training, while keeping the finetuning temporal stride fixed to the value $\tau=8$ (i.e., sampling a frame every 8 from the video starting from a random frame). Figure~\ref{fig:sampling rate} shows how different combinations of $\tau_S$ and $\tau_L$ affect the final video-level accuracy on Kinetics-400. For ease of interpretation we split the visualization of results over 4 distinct plots, representing 4 different values of $\tau_S$: $\tau_S \in \{4, 8, 16, 32\}$. Each plot shows how the final video-level accuracy varies for different temporal stride values $\tau_L$ of the long clip where $\tau_L \geq \tau_S$ and $\tau_S$ is kept fixed. There are two important observations we can make from these results. The first is that, for each choice of $\tau_S$, the larger the gap between the two strides (i.e., the larger the value of $\tau_L - \tau_S$), the higher is the accuracy. This can be seen in the first three plots where the accuracy curve monotonically increases as $\tau_L$ is made larger starting from the initial value of $\tau_L=\tau_S$. This validates the importance of contrasting views of different temporal lengths during self-supervised pretraining. The second observation is that our model performs best when $\tau_S=8$ and $\tau_L=32$. This result makes intuitive sense as a short clip sampled with $\tau_S=8$ is temporally long enough to allow to predict the context of the long clip; at the same time it is short enough to allow the method to use a long view that is significantly longer (up to 4 times longer than the short view). Conversely, choosing a larger value of $\tau_S$ (i.e., 16 or 32) reduces the maximum possible gap $\tau_L - \tau_S$ between the two views, while choosing a smaller value of $\tau_S$ (i.e., 4) would cause the contrastive learning between the two views to be overly difficult due to the excessive brevity of the short clip. 

% Figure~\ref{fig:time_scale} aggregates the four curves corresponding to $\tau_S \in \{4, 8, 16, 32\}$ into a single plot in order to compare them more easily. Furthermore, in this plot we include additional performance points corresponding to three new settings: 1) the middle red triangle in Figure~\ref{fig:time_scale} depicts the performance of our system when $\tau_L=32$ and when $\tau_S$ is sampled randomly from an interval  $\{4,8,16\}$ for each short clip; 2) the right-most blue circle represents the opposite setting where $\tau_S$ is kept fixed ($\tau_S=4$) and $\tau_{L}$ is randomly sampled from an interval $\{8,16,32\}$; 3) the right-most red triangle shows a configuration where both temporal strides are randomly chosen for each video clip from the following values $\{4,8,16,32\}$. The results in~\ref{fig:time_scale} clearly show that adding randomness in the choice of the temporal extents for the long and short clips does not produce improved performance. The best performance is still achieved when $\tau_S=8$ and $\tau_L=32$. Thus, we adopt this setup for all subsequent experiments.

\begin{table}[]
    \centering
    \footnotesize
    \begin{tabular}{c c c}
        \hline
        $\tau_S$ & $\tau_L$            & Accuracy \\ \hline
        4 & \{8,16,32\}  &  73.9         \\
         \{4,8,16\} & 32  &  74.8         \\
         8 & \{8, 16,32\} & 75.5\\
         \{8,16,32\} &32 & 75.9\\
         \{4,8,16,32\}& \{4,8,16,32\}  &  75.6         \\
         \{8,16,32\}& \{8,16,32\}  &  76.0      \\
         8 & 32        &    \bf 76.6       \\ \hline
    \end{tabular}
    \caption{We analyze the potential benefits of randomly sampling either $\tau_S$ and/or $\tau_L$ (for the short and the long clips, respectively). Accuracy is measured for video-level classification on Kinetics-400 after pretraining with our LSTCL system using MoCo v3. The best result is still achieved for fixed values of $\tau_S=8$ and $\tau_L=32$.}
    \label{tab:time_scale}
\end{table}

In Table~\ref{tab:time_scale}, we include additional performance points corresponding to settings where $\tau_S$ and/or $\tau_L$ are sampled randomly for each training video clip. Specifically, the first row in the table shows the performance of our system when $\tau_S=4$ and $\tau_L$ is sampled randomly from $\{8,16,32\}$; the second row represents the opposite setting where $\tau_L$ is kept fixed ($\tau_L=32$) and $\tau_{S}$ is randomly sampled from $\{4, 8,16\}$; the setting in the third row is similar to that of the first row but with $\tau_S=8$; the fourth row shows the same setting as the second row but excludes $\tau_S=4$; the fifth and the sixth rows show configurations where both temporal strides are randomly chosen for each training video clip, subject to $\tau_S \leq \tau_L$. As before, we keep the finetuning temporal stride fixed to the value $\tau=8$. The results in Table~\ref{tab:time_scale} clearly show that adding randomness in the choice of the temporal extents for the long and short clips does not produce improved performance. The best performance is still achieved when $\tau_S=8$ and $\tau_L=32$ (shown in the last row). Thus, we adopt this setup for all subsequent experiments.

\begin{figure}[]
\centering
\begin{subfigure}{.235\textwidth}
  \includegraphics[width=1.0\linewidth]{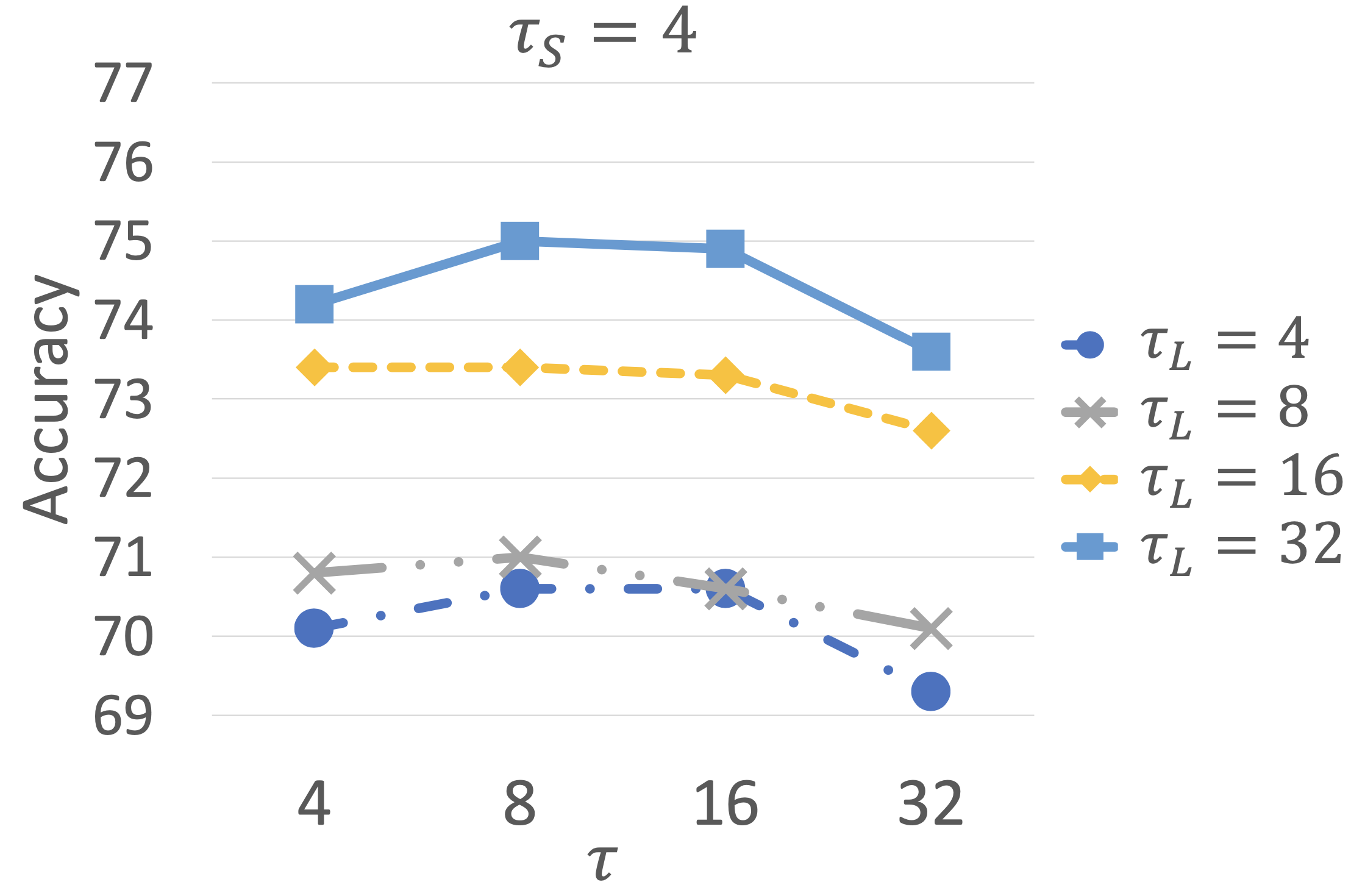}
\end{subfigure}
\begin{subfigure}{.235\textwidth}
  \includegraphics[width=1.0\linewidth]{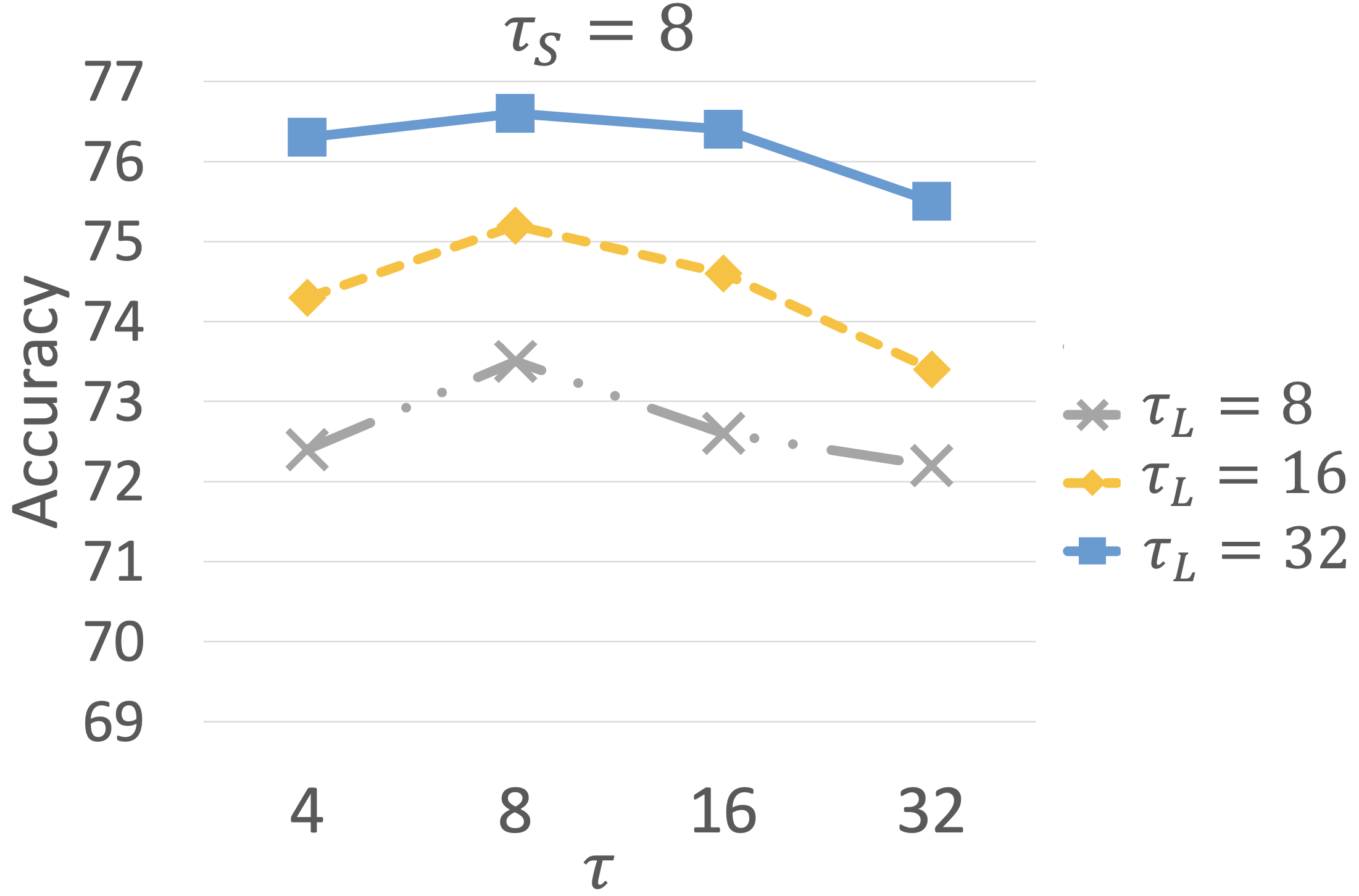}
\end{subfigure}
\caption{The plots show video-level accuracy on Kinetics-400 for different values of the temporal sampling stride $\tau$ used for supervised finetuning and test-time inference.}
\label{fig:inferenceSR}
\end{figure}

Now we turn to study the impact of the {\em finetuning} stride $\tau$ on video-level accuracy. The two plots in Figure~\ref{fig:inferenceSR} show how the accuracy on Kinetics-400 varies as we change the value of $\tau$ (on the horizontal axis) for two different choices of $\tau_S$ ($\tau_S=4$ in the left plot, and $\tau_S=8$ in the right plot). The different curves in each plot correspond to different choices of $\tau_L$. We see that setting the finetuning stride to $\tau=8$ tends to produce the best results across all possible choices of $\tau_S$ and $\tau_L$. This makes sense, since with $\tau=8$  the 5 inference clips are short enough not to overlap so that they provide complementary information for the video-level classification. At the same time, $\tau=8$ implies an inference clip long enough to yield good classification on its own.

\begin{figure}
    \centering
    \includegraphics[width=0.7\linewidth]{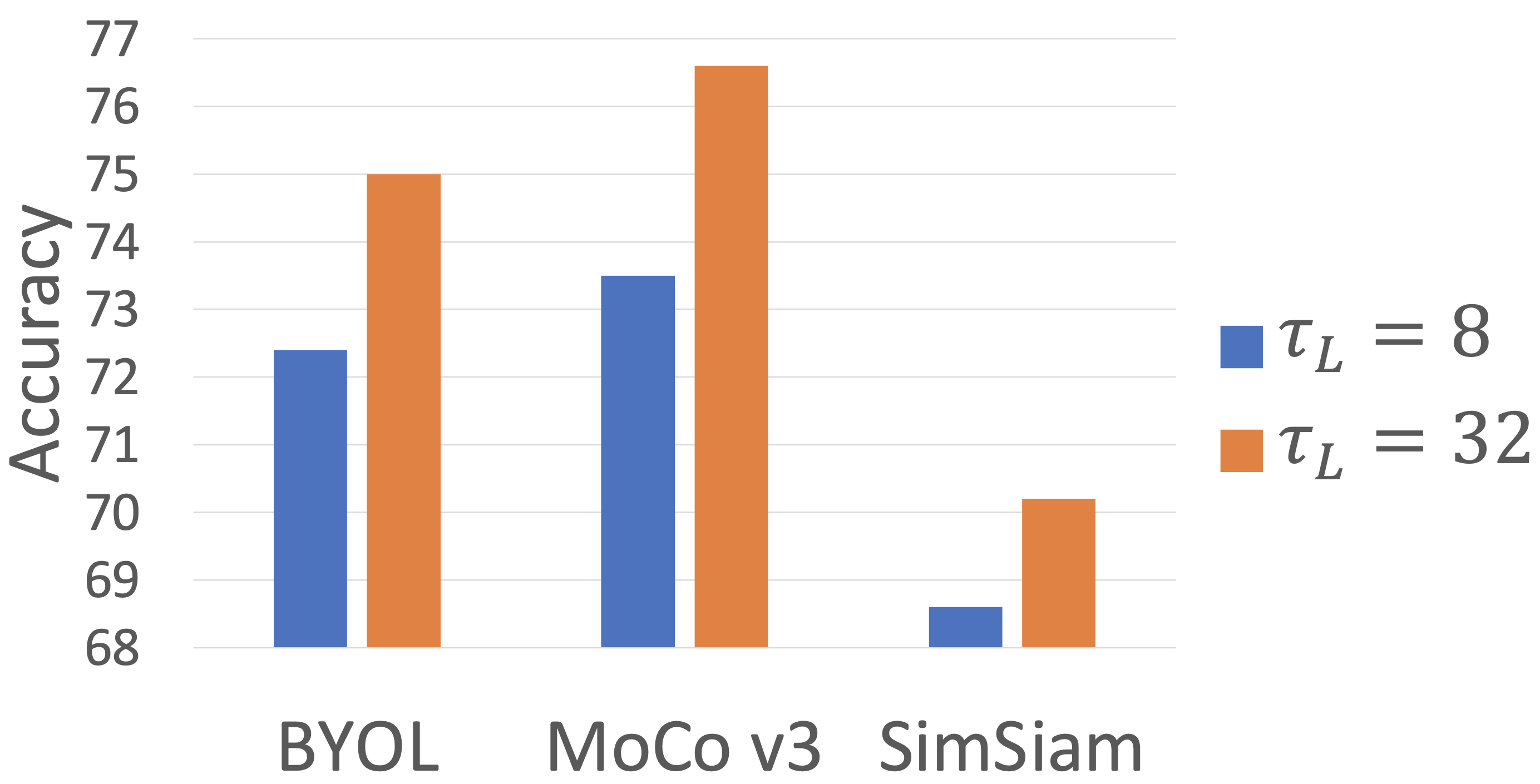}
    \caption{Kinetics-400 accuracy achieved by pretraining with LSTCL using three self-supervised strategies, with two possible stride values for the long clip ($\tau_L\in\{8, 32\})$ (stride for the short clip is fixed to $\tau_S=8$). All three methods benefit from using views of different lengths ($\tau_L=32$, instead of $\tau_L=\tau_S =8$).}
    \label{fig:TL}
\end{figure}

\noindent\textbf{Different Contrastive Learning Frameworks.} Next, we investigate the effects of different contrastive learning frameworks in our LSTCL system. Specifically, we experiment with three recent approaches: BYOL, MoCo v3, and SimSiam. Figure~\ref{fig:TL} shows that a larger temporal stride $\tau_L$ for the long view leads to better accuracy for all three of these frameworks. Specifically, setting $\tau_L=32$ leads to the following performance gains compared to the setting where $\tau_L=\tau_S=8$: +2.6\% for BYOL, +3.1\% for MoCo v3, and +1.6\% for SimSiam. The lower absolute performance of SimSiam can be explained by the lack of the momentum-encoder, which we observed to be important when training video transformer models with LSTCL. Thus, based on these result, for all subsequent experiments, we adopt MoCo v3 as our base learning framework.

\textbf{Weight Sharing and Contrastive Loss.} Here we ablate the two main differences between LSTCL and BraVe~\cite{recasens2021broaden}. 1) BraVe has two independent backbones, projectors and predictors, which define a broad stream and a narrow stream. Instead, our LSTCL adopts online and momentum encoders with shared parameters. 2) Each stream in BraVe is specialized to process a particular type of view (either broad or narrow). Training is done by means of a combination of two regression objectives (one mapping from broad to narrow, the other mapping in the opposite direction). In LSTCL, a single encoder takes both views. Our model is optimized using a single contrastive loss, which minimizes differences between the two views. 

In Table~\ref{weight_loss}, we present ablation results of LSTCL with respect to differences 1) and 2) outlined above. For 1), we modify LSTCL to use distinct networks (independent backbones and projectors) for the two views, as in BraVe. 2) In addition to using separate networks, we adopt the data feeding and learning objectives from BraVe in our LSTCL. From the results, it can be seen that LSTCL (first row) achieves superior performance with only half the number of parameters compared to these two alternative setups.

%In LSTCL, both the backbone and the projector of the two encoders share weights through a momentum factor and the final objective is formulated through an InfoNCE loss between the final representations of the long and the short clip. Instead, in BraVe~\cite{recasens2021broaden} the backbone in each stream is specialized to process a particular type of view (either broad or narrow). Training is done by means of a combination of two regression objectives (one mapping from broad to narrow, the other mapping in the opposite direction). In Table~\ref{weight_loss}, we study the effect of these two alternative designs in our LSTCL. From the results, it can be seen that our default architecture achieves superior performance with only half the number of parameters compared to the two alternative setup. 

\begin{table}[]
\centering
\footnotesize
\begin{tabular}{l c c c}
\hline
Loss  & Shared Backbone & Accuracy & Params \\\hline
InfoNCE    & Yes             & \bf 76.6     & \bf 121.4M     \\
InfoNCE    & No              & 73.2     & 242.8M     \\
Regression & No              & 70.8     & 242.8M     \\\hline
\end{tabular}
\caption{We compare our proposed approach (first row) against the weight sharing and loss proposed in BraVe~\cite{recasens2021broaden} by evaluating the effects on Kinetics-400.}
\label{weight_loss}
\end{table}

% \begin{figure}[t]
% \begin{center}
%  fig:time_scale
% \end{center}
%   \caption{Performance of LSTCL when using different self-supervised learning frameworks. Using a larger frame sampling rate $\tau_L$ leads to a consistently higher accuracy on Kinetics-400.}
% \label{fig:TL}
% \end{figure}
\begin{table}[]
\footnotesize
    \centering
    \begin{tabular}{l c}
        \hline
        Sampling Method               & Accuracy \\ \hline
        Random Disjoint               & 72.6          \\
        Random Included               & 76.2  \\
        Random Independent            & \bf 76.6 \\ \hline
        \end{tabular}
    \caption{Comparison of different clip sampling strategies for LSTCL on Kinetics-400. In these experiments we use $\tau_S=8$ and $\tau_L=32$ for LSTCL and $\tau=8$ for finetuning.}
    \label{Sample}
\end{table}

\noindent\textbf{Clip Sampling Strategy in LSTCL.}
In Table~\ref{Sample}, we study the effect of different clip sampling strategies. These results indicate that random independent sampling works best in our setting. Intuitively, this makes sense as it forces our model to extrapolate to arbitrary video views.

\noindent\textbf{Video Transformers.}
In Table~\ref{Finetuning} we compare the performance of three distinct video transformer architectures: TimeSformer, Swin, and Space-Time (ST) Swin. We train each of these models under three different scenarios on Kinetics-400: 1) from scratch (without pretraining), 2) using supervised pretraining on the large-scale ImageNet-1K dataset, and 3) using our self-supervised LSTCL pretraining. We can see that among these three training strategies, our LSTCL pretraining provides the highest accuracy, outperforming the models that use large-scale supervised ImageNet-1K pretraining for all three architectures.

\noindent\subsection{Comparison to the State-of-the-Art}
\label{soa}

For our final experiments, we adopt the Space-Time Swin transformer as it achieves the strongest results in our ablation studies. For this comparison to the state-of-the-art, we also train models using clips of $T=16$ frames during both pretraining with LSTCL and supervised finetuning. Even in this case we set the temporal stride to $\tau_S=4$ for short clips and $\tau_L=16$ for long clips.

\noindent\textbf{Kinetics-400 \& Kinetics-600.} In Table~\ref{k400}, we report results on Kinetics-400, listing for each method the clip size, the accuracy, the  inference cost (in TFLOPs), and the number of parameters. We group methods on the basis of the input clip size, since models trained on longer clips or higher resolution frames tend to yield higher accuracy. The first two groups include models operating on clips of the same size as those used by our system ($8 \times 224^2$ and $16 \times 224^2$). It can be seen that the ST Swin model pretrained with LSTCL achieves the highest accuracy among all previous methods that use the same input clip size and that do not make use of additional data. Furthermore, compared to prior video transformer models that are  pretrained with full supervision on large-scaled labeled datasets (the bottom part of the table), our method still achieves competitive results and actually often yields better accuracy. Lastly, note that compared to training our ST Swin model from scratch, LSTCL pretraining leads to a significant $8.7\%$ boost on Kinetics-400.

\begin{table}[]
\footnotesize
    \centering
    \begin{tabular}{c c c c c}
        \hline
        Model    & Scratch & IN-1K  &LSTCL & Params\\ \hline
        TimeSformer~\cite{bertasius2021space}  & 60.4     &75.8          & \bf 76.6  &121.4M          \\
        Swin         & 66.2     &73.3     & \bf 75.5    & 88.0M        \\ 
        ST Swin      & 71.1     &76.0     & \bf 79.8    & 88.0M         \\ \hline
        \end{tabular}
        \vspace{-0.2cm}
    \caption{Comparing self-supervised pretraining using LSTCL to training from scratch and supervised pretraining on ImageNet-1K (IN-1K). The results show video classification accuracy on Kinetics-400 for three video transformer architectures.\vspace{-0.3cm}}
    \label{Finetuning}
\end{table}

% \begin{table}[!ht]
% \footnotesize
% \centering
% \begin{tabular}{|l|c|c|c|c|}
% \hline
% Model       & Pretraining dataset & Finetuning dataset & Acc. \\ \hline \hline
% ST Swin  &IN-1K (Superv.) & HMDB51 & 40.2\% \\ \hline
% ST Swin  &K400 (Superv.) & HMDB51 & 61.2\% \\ \hline 
% ST Swin  &K400 (LSTCL) & HMDB51 & 75.9\% \\ \hline \hline
% ST Swin  &IN-1K (Superv.) & UCF101 & 78.1\% \\ \hline
% ST Swin  &K400 (Superv.) & UCF101 & 88.9\% \\ \hline
% ST Swin  &K400 (LSTCL) & UCF101 & 96.8\% \\ \hline
% \end{tabular}
% \caption{Comparing LSTCL vs supervised pretraining (on IN-1K and K400) for video classification in HMDB51 and UCF101.}
% \label{transferlearning}
% \end{table}

\begin{table}[!ht]
\footnotesize
\centering
\begin{tabular}{lcccc}
\hline
Model       & Pretraining dataset & UCF101 & HMDB51 \\ \hline
BraVe~\cite{recasens2021broaden} & K400 (Unsup.) & 95.1 & 74.6\\
$\rho$BYOL~\cite{Feichtenhofer_large} & K400 (Unsup.) & 96.3 & 75.0\\\hline
ST Swin  &IN-1K (Superv.) & 78.1 & 40.2 \\ 
ST Swin  &K400 (Superv.) & 88.9 & 61.2 \\ 
\bf ST Swin w/ LSTCL  &K400 (Unsup.) & \bf 96.8 &  \bf 75.9\\ \hline
\end{tabular}
\vspace{-0.2cm}
\caption{Transfer learning results on UCF101 and HMDB51. We report the performance using the full fine-tuning setting. Our method outperforms previous state-of-the-art approaches on both UCF101 and HMDB51. Furthermore, our unsupervised LSTCL pretraining scheme achieves better results than the approaches based on supervised pretraining (on IN-1K and K400)\vspace{-0.3cm}.}
\label{transferlearning}
\end{table}

Table~\ref{k600} shows a comparison with the state-of-the-art on the Kinetics-600 dataset. Even here we see that ST Swin pretrained with LSTCL achieves the best accuracy within the two groups of models using the same clip sizes as our networks. Furthermore, LSTCL produces a gain of $7.3\%$ compared to learning from scratch.

\noindent\textbf{Something-Something-V2.} In  Table~\ref{ssv2} we report the performance on the Something-Something-V2 dataset. Most prior methods leverage {\em supervised} large-scale pre-training on external datasets in order to achieve strong performance on this benchmark, since the dataset is relatively small. The results in the table highlight that our ST Swin models pretrained {\em without} labels on Kinetics-400 using LSTCL achieves higher accuracy than methods that leverage pretraining on larger datasets and using manually labeled data. Moreover, our LSTCL pretraining yields a gain of $26.4\%$ over the same model trained from scratch. This remarkable improvement is due to the fact that the Something-Something-V2 dataset requires thorough temporal reasoning for good accuracy. Our LSTCL method trains the clip representation to predict the temporal context from the entire video and thus yields strong benefits on this benchmark.

\begin{table*}[!htb]

\centering
\caption{Comparison to the state-of-the-art on Kinetics-400. Among methods using the same clip sizes as our models and no additional data (first two groups in the table), ST Swin networks pretrained with LSTCL achieve the highest accuracy and they are are on-par with models that use longer or higher-resolution clips (third group) or that leverage additional data for supervised pretraining (bottom group).}
{\footnotesize
\begin{tabular}{c c c c c c c }
\hline
Method          & Clip Size & Additional Data  (\# Samples) & Top-1 & Top-5 & TFLOPs & Params   \\ \hline
SlowFast~\cite{feichtenhofer2019slowfast}&$8\ \mathrm{x}\ 224^2$ & - &77.9 & 93.2 & 3.0 &59.9M\\
TimeSformer-scratch   &$8\ \mathrm{x}\ 224^2$  & -   & 60.4 & 76.7    &  0.59 &121.4M  \\ 
ST Swin from scratch &$8\ \mathrm{x}\ 224^2$ & -  & 71.1 & 85.2 & 0.60 &88.0M\\
\bf ST Swin w/ LSTCL &$8\ \mathrm{x}\ 224^2$ & -  & \bf 79.8 & \bf 94.0 & 0.60 &88.0M\\ \hline
CorrNet-101~\cite{wang2020video} &$16\ \mathrm{x}\ 224^2$ & - &79.2 & - & 7.0 &-\\
SlowFast-NL~\cite{feichtenhofer2019slowfast}&$16\ \mathrm{x}\ 224^2$ & - &79.8 & 93.9 & 7.0 &59.9M\\
MViT-B ~\cite{fan2021multiscale}    &$16\ \mathrm{x}\ 224^2$ & -         & 78.4      & 93.5 & 0.36 &36.6M \\
\bf ST Swin w/ LSTCL &$16\ \mathrm{x}\ 224^2$ & -  & \bf 81.5 & \bf 95.2 & 1.80 &88.0M\\\hline
X3D-XL~\cite{feichtenhofer2020x3d} & $16\ \mathrm{x}\ 312^2$ & - &79.1 &93.9 &1.45 &11.0M\\
ip-CSN-152 ~\cite{tran2019video}&$32\ \mathrm{x}\ 224^2$ & - &77.8 & 92.8 & 3.3 &32.8M\\
MViT-B ~\cite{fan2021multiscale}    &$64\ \mathrm{x}\ 224^2$ & -         & 81.2      & 95.1 & 4.09 &36.6M \\ \hline
TimeSformer~\cite{bertasius2021space}   &$8\ \mathrm{x}\ 224^2$  & {ImageNet-21K}  (14M) & 78.0 &93.7    &  0.59 &121.4M  \\ 
STAM~\cite{sharir2021image}    &$16\ \mathrm{x}\ 224^2$ &   ImageNet-21K (14M) & 79.3      & - & 0.27 &96.0M \\ 
TEINet~\cite{liu2020teinet}    &$16\ \mathrm{x}\ 224^2$ &   ImageNet-1K (1.2M) & 76.2      & 92.5 & 1.8 & - \\ 
Mformer~\cite{patrick2021keeping} &$16\ \mathrm{x}\ 224^2$ & ImageNet-21K (14M) & 79.7 & 94.2 & 11.1 & 109.1M \\
ViViT-L~\cite{arnab2021vivit}    &$16\ \mathrm{x}\ 224^2$ &   ImageNet-21K (14M) & 80.6      & 94.7 & 47.9 &310.0M \\ 
VATT-B~\cite{akbari2021vatt} &$32\ \mathrm{x}\ 320^2$ &  AudioSet + HowTo100M (3.2M) & 79.6 & 94.9 & 9.08 & 88.0M \\
TimeSformer-L~\cite{bertasius2021space} &$96\ \mathrm{x}\ 224^2$ &  ImageNet-21K (14M)  & 80.7  & 94.7  &  7.14 &121.4M \\ \hline
% Ours (ST Swin) &16 x 224^2 & Self-Supervised & xx.x&xx.x &246 x 5 x 1 &88.0M\\\hline
\end{tabular}
}
\label{k400}
\end{table*}

\begin{table*}[!ht]
\centering
\caption{Video-level accuracy on Kinetics-600. The ST Swin model trained with LSTCL achieves results comparable with the state-of-the-art {\em without} using additional data or labels.}
{\footnotesize
\begin{tabular}{ c c c c c }
\hline
Method          & Clip Size & Additional Data (\# Samples) & Top-1 & Top-5    \\ \hline
SlowFast~\cite{feichtenhofer2019slowfast}&$8\ \mathrm{x}\ 224^2$ & - &80.4 & 94.8 \\
 ST Swin from scratch &$8\ \mathrm{x}\ 224^2$ & -  &74.7  &92.2   \\
\bf ST Swin w/ LSTCL &$8\ \mathrm{x}\ 224^2$ & - &\bf 82.0 &\bf 95.5  \\ \hline
SlowFast~\cite{feichtenhofer2019slowfast}&$16\ \mathrm{x}\ 224^2$ & - &81.8 & 95.1 \\
MViT-B~\cite{fan2021multiscale}&$16\ \mathrm{x}\ 224^2$    & -        & 82.1      & 95.7   \\ 
\bf ST Swin w/ LSTCL &$16\ \mathrm{x}\ 224^2$ & - &\bf 83.6 & \bf 96.6  \\\hline
X3D-XL~\cite{feichtenhofer2020x3d} & $16\ \mathrm{x}\ 312^2$ & - &81.9 &95.9\\
MViT-B~\cite{fan2021multiscale}&$32\ \mathrm{x}\ 224^2$    & -         & 83.4     & 96.3   \\ \hline
TimeSformer~\cite{bertasius2021space} &$8\ \mathrm{x}\ 224^2$    & {ImageNet-21K}  (14M) & 79.1  &94.4     \\ 
Mformer~\cite{patrick2021keeping} &$16\ \mathrm{x}\ 224^2$   &{ImageNet-21K}  (14M) &81.6 &95.6\\ 
ViViT-L~\cite{arnab2021vivit} &$16\ \mathrm{x}\ 224^2$   & {ImageNet-21K}  (14M) & 82.5      & 95.6  \\ 
VATT-B~\cite{akbari2021vatt} &$32\ \mathrm{x}\ 320^2$ &   AudioSet + HowTo100M (3.2M) & 80.5 & 95.5 \\
VATT-L~\cite{akbari2021vatt} &$32\ \mathrm{x}\ 320^2$ &   AudioSet + HowTo100M (3.2M) & 83.6 & 96.6  \\
TimeSformer-L~\cite{bertasius2021space}&$96\ \mathrm{x}\ 224^2$  & ImageNet-21K  (14M)  & 82.2  & 95.6   \\ \hline
% Ours (ST Swin) &16 x 224^2 & \makecell{K400\\Self-Supervised} & &  \\\hline
\end{tabular}
}
\label{k600}
\end{table*}

\begin{table*}[!ht]
\centering
\caption{Video-level classification accuracy on Something-Something-V2. Our ST Swin models pretrained {\em without} labels
using LSTCL yield results on par with the state-of-the-art.}
{\footnotesize
\begin{tabular}{ c c c c c c }
\hline
Method          & Clip Size & Additional Data  (\# Samples)  & Pretraining & Top-1 & Top-5   \\ \hline
TimeSformer~\cite{bertasius2021space}&$8\ \mathrm{x}\ 224^2$   & {ImageNet-21K}  (14M) & supervised & 59.5  &-     \\ 
ResNet50~\cite{Feichtenhofer_large} &$8\ \mathrm{x}\ 224^2$ &  K400 (240K) & unsupervised  &55.8 & - \\
ST Swin from scratch &$8\ \mathrm{x}\ 224^2$ &  - & -    &38.4  &65.5   \\
\bf ST Swin w/ LSTCL &$8\ \mathrm{x}\ 224^2$ &  K400 (240K) & unsupervised    &\bf 64.8  &\bf 89.4  \\\hline
TEINet~\cite{liu2020teinet}    &$16\ \mathrm{x}\ 224^2$ &   ImageNet-1K (1.2M) & supervised      & 64.7 & -  \\ 
Mformer~\cite{patrick2021keeping} &$16\ \mathrm{x}\ 224^2$ &   ImageNet-21K + K400 (14.2M) & supervised      & 66.5 & 90.1  \\ 
ViViT-L~\cite{arnab2021vivit} & $16\ \mathrm{x}\ 224^2$    & {ImageNet-21K}  (14M) & supervised & 65.4 & 89.8 \\
MViT-B~\cite{fan2021multiscale}&$16\ \mathrm{x}\ 224^2$    & K400 (240K) & supervised       & 64.7 &89.2      \\ 
\bf ST Swin w/ LSTCL &$16\ \mathrm{x}\ 224^2$ &  K400 (240K) & unsupervised    &\bf 67.0 &\bf 90.5  \\\hline
TimeSformer-L~\cite{bertasius2021space}&$96\ \mathrm{x}\ 224^2$ & {ImageNet-21K}  (14M) & supervised & 62.4  &- \\ 
MViT-B~\cite{fan2021multiscale}&$64\ \mathrm{x}\ 224^2$    & K400 (240K) & supervised         & 67.7     & 90.9   \\  \hline
% Ours (ST Swin)&16 x 224^2 &\makecell{K400\\Self-Supervised} & &  \\\hline
\end{tabular}
}
\label{ssv2}
\end{table*}

\noindent\textbf{HMDB51 \& UCF101.} Finally, we assess the ability to transfer the unsupervised representation learned by LSTCL from Kinetics-400 to the small-scale datasets of HMDB~\cite{Kuehne11} and UCF101~\cite{UCF101} via supervised finetuning. The results are shown in Table~\ref{transferlearning} where we include also accuracies obtained via fully-supervised pretraining (using class labels) on IN-1K and K400 and also the two recent self-supervised methods $\rho$BYOL~\cite{Feichtenhofer_large} and BraVe~\cite{recasens2021broaden}. It can be seen that LSTCL ourperforms both (i) the previous state-of-the-art unsupervised pretraining methods, and (ii) the supervised pretraining baselines on both datasets. %This suggests that LSTCL represents an effective option for transfer learning to novel classes and datasets, yielding superior results to the standard practice of supervised pretraining.

\section{Conclusion}
\label{conclusion}
This paper introduces Long-Short Temporal Contrastive Learning (LSTCL), an unsupervised pretraining scheme for video transformers. By contrasting representations obtained from a long view and a short view of each video, it forces the model to encode context from the whole video into the features of short clips. We demonstrate our LSTCL under three different contrastive frameworks and two video transformer architectures including a new variant, Space-Time Swin transformer. %LSTCL eliminates the need for large-scaled supervised image pretraining in video transformers. 
In our experiments we show that unsupervised pretraining with LSTCL leads to similar or better video classification accuracy compared to pretraining with full supervision on ImageNet-21K and it achieves competitive results on three different video classification benchmarks. LSTCL effectively eliminates the need for large-scaled supervised image pretraining in video transformers. 
%and LSTCL with Space-Time Swin transformer also achieves strong performance in multiple benchmarks. We hope our work will provide more intuition for training video transformers additional-data-free.

%%%%%%%%% REFERENCES
{\small
\bibliographystyle{ieee_fullname}
\bibliography{egbib}

\begin{thebibliography}{10}\itemsep=-1pt

\bibitem{Agrawal_2015_ICCV}
Pulkit Agrawal, Joao Carreira, and Jitendra Malik.
\newblock Learning to see by moving.
\newblock In {\em Proceedings of the IEEE International Conference on Computer
  Vision (ICCV)}, December 2015.

\bibitem{akbari2021vatt}
Hassan Akbari, Linagzhe Yuan, Rui Qian, Wei-Hong Chuang, Shih-Fu Chang, Yin
  Cui, and Boqing Gong.
\newblock Vatt: Transformers for multimodal self-supervised learning from raw
  video, audio and text.
\newblock {\em arXiv preprint arXiv:2104.11178}, 2021.

\bibitem{arnab2021vivit}
Anurag Arnab, Mostafa Dehghani, Georg Heigold, Chen Sun, Mario Lu\v{c}i\'c, and
  Cordelia Schmid.
\newblock Vivit: A video vision transformer.
\newblock In {\em Proceedings of the IEEE/CVF International Conference on
  Computer Vision (ICCV)}, pages 6836--6846, October 2021.

\bibitem{ba2016layer}
Jimmy~Lei Ba, Jamie~Ryan Kiros, and Geoffrey~E Hinton.
\newblock Layer normalization.
\newblock {\em arXiv preprint arXiv:1607.06450}, 2016.

\bibitem{benaim2020speednet}
Sagie Benaim, Ariel Ephrat, Oran Lang, Inbar Mosseri, William~T Freeman,
  Michael Rubinstein, Michal Irani, and Tali Dekel.
\newblock Speednet: Learning the speediness in videos.
\newblock In {\em Proceedings of the IEEE/CVF Conference on Computer Vision and
  Pattern Recognition}, pages 9922--9931, 2020.

\bibitem{bertasius2021space}
Gedas Bertasius, Heng Wang, and Lorenzo Torresani.
\newblock Is space-time attention all you need for video understanding?
\newblock In Marina Meila and Tong Zhang, editors, {\em Proceedings of the 38th
  International Conference on Machine Learning, {ICML} 2021, 18-24 July 2021,
  Virtual Event}, volume 139 of {\em Proceedings of Machine Learning Research},
  pages 813--824. {PMLR}, 2021.

\bibitem{carion2020end}
Nicolas Carion, Francisco Massa, Gabriel Synnaeve, Nicolas Usunier, Alexander
  Kirillov, and Sergey Zagoruyko.
\newblock End-to-end object detection with transformers.
\newblock In {\em European Conference on Computer Vision}, pages 213--229.
  Springer, 2020.

\bibitem{caron2020unsupervised}
Mathilde Caron, Ishan Misra, Julien Mairal, Priya Goyal, Piotr Bojanowski, and
  Armand Joulin.
\newblock Unsupervised learning of visual features by contrasting cluster
  assignments.
\newblock {\em arXiv preprint arXiv:2006.09882}, 2020.

\bibitem{caron2021emerging}
Mathilde Caron, Hugo Touvron, Ishan Misra, Herv{\'e} J{\'e}gou, Julien Mairal,
  Piotr Bojanowski, and Armand Joulin.
\newblock Emerging properties in self-supervised vision transformers.
\newblock {\em arXiv preprint arXiv:2104.14294}, 2021.

\bibitem{carreira2017quo}
Joao Carreira and Andrew Zisserman.
\newblock Quo vadis, action recognition? a new model and the kinetics dataset.
\newblock In {\em proceedings of the IEEE Conference on Computer Vision and
  Pattern Recognition}, pages 6299--6308, 2017.

\bibitem{chen2020simple}
Ting Chen, Simon Kornblith, Mohammad Norouzi, and Geoffrey Hinton.
\newblock A simple framework for contrastive learning of visual
  representations.
\newblock In {\em International conference on machine learning}, pages
  1597--1607. PMLR, 2020.

\bibitem{chen2020improved}
Xinlei Chen, Haoqi Fan, Ross Girshick, and Kaiming He.
\newblock Improved baselines with momentum contrastive learning.
\newblock {\em arXiv preprint arXiv:2003.04297}, 2020.

\bibitem{chen2020exploring}
Xinlei Chen and Kaiming He.
\newblock Exploring simple siamese representation learning.
\newblock {\em arXiv preprint arXiv:2011.10566}, 2020.

\bibitem{chen2021empirical}
Xinlei Chen, Saining Xie, and Kaiming He.
\newblock An empirical study of training self-supervised visual transformers.
\newblock {\em arXiv preprint arXiv:2104.02057}, 2021.

\bibitem{devlin2018bert}
Jacob Devlin, Ming-Wei Chang, Kenton Lee, and Kristina Toutanova.
\newblock Bert: Pre-training of deep bidirectional transformers for language
  understanding.
\newblock {\em arXiv preprint arXiv:1810.04805}, 2018.

\bibitem{dosovitskiy2020image}
Alexey Dosovitskiy, Lucas Beyer, Alexander Kolesnikov, Dirk Weissenborn,
  Xiaohua Zhai, Thomas Unterthiner, Mostafa Dehghani, Matthias Minderer, Georg
  Heigold, Sylvain Gelly, et~al.
\newblock An image is worth 16x16 words: Transformers for image recognition at
  scale.
\newblock {\em arXiv preprint arXiv:2010.11929}, 2020.

\bibitem{fan2021multiscale}
Haoqi Fan, Bo Xiong, Karttikeya Mangalam, Yanghao Li, Zhicheng Yan, Jitendra
  Malik, and Christoph Feichtenhofer.
\newblock Multiscale vision transformers.
\newblock In {\em Proceedings of the IEEE/CVF International Conference on
  Computer Vision (ICCV)}, pages 6824--6835, October 2021.

\bibitem{feichtenhofer2020x3d}
Christoph Feichtenhofer.
\newblock X3d: Expanding architectures for efficient video recognition.
\newblock In {\em Proceedings of the IEEE/CVF Conference on Computer Vision and
  Pattern Recognition}, pages 203--213, 2020.

\bibitem{feichtenhofer2019slowfast}
Christoph Feichtenhofer, Haoqi Fan, Jitendra Malik, and Kaiming He.
\newblock Slowfast networks for video recognition.
\newblock In {\em Proceedings of the IEEE/CVF International Conference on
  Computer Vision}, pages 6202--6211, 2019.

\bibitem{Feichtenhofer_large}
Christoph Feichtenhofer, Haoqi Fan, Bo Xiong, Ross~B. Girshick, and Kaiming He.
\newblock A large-scale study on unsupervised spatiotemporal representation
  learning.
\newblock In {\em {IEEE} Conference on Computer Vision and Pattern Recognition,
  {CVPR} 2021, virtual, June 19-25, 2021}, pages 3299--3309. Computer Vision
  Foundation / {IEEE}, 2021.

\bibitem{goodfellow2016deep}
Ian Goodfellow, Yoshua Bengio, Aaron Courville, and Yoshua Bengio.
\newblock {\em Deep learning}, volume~1.
\newblock MIT press Cambridge, 2016.

\bibitem{goodfellow2014generative}
Ian~J Goodfellow, Jean Pouget-Abadie, Mehdi Mirza, Bing Xu, David Warde-Farley,
  Sherjil Ozair, Aaron Courville, and Yoshua Bengio.
\newblock Generative adversarial networks.
\newblock {\em arXiv preprint arXiv:1406.2661}, 2014.

\bibitem{goroshin2015unsupervised}
Ross Goroshin, Joan Bruna, Jonathan Tompson, David Eigen, and Yann LeCun.
\newblock Unsupervised learning of spatiotemporally coherent metrics.
\newblock In {\em Proceedings of the IEEE international conference on computer
  vision}, pages 4086--4093, 2015.

\bibitem{goyal2017accurate}
Priya Goyal, Piotr Doll{\'a}r, Ross Girshick, Pieter Noordhuis, Lukasz
  Wesolowski, Aapo Kyrola, Andrew Tulloch, Yangqing Jia, and Kaiming He.
\newblock Accurate, large minibatch sgd: Training imagenet in 1 hour.
\newblock {\em arXiv preprint arXiv:1706.02677}, 2017.

\bibitem{goyal2017something}
Raghav Goyal, Samira Ebrahimi~Kahou, Vincent Michalski, Joanna Materzynska,
  Susanne Westphal, Heuna Kim, Valentin Haenel, Ingo Fruend, Peter Yianilos,
  Moritz Mueller-Freitag, et~al.
\newblock The" something something" video database for learning and evaluating
  visual common sense.
\newblock In {\em Proceedings of the IEEE International Conference on Computer
  Vision}, pages 5842--5850, 2017.

\bibitem{grill2020bootstrap}
Jean-Bastien Grill, Florian Strub, Florent Altch{\'e}, Corentin Tallec,
  Pierre~H Richemond, Elena Buchatskaya, Carl Doersch, Bernardo~Avila Pires,
  Zhaohan~Daniel Guo, Mohammad~Gheshlaghi Azar, et~al.
\newblock Bootstrap your own latent: A new approach to self-supervised
  learning.
\newblock {\em arXiv preprint arXiv:2006.07733}, 2020.

\bibitem{han2019video}
Tengda Han, Weidi Xie, and Andrew Zisserman.
\newblock Video representation learning by dense predictive coding.
\newblock In {\em Proceedings of the IEEE/CVF International Conference on
  Computer Vision Workshops}, pages 0--0, 2019.

\bibitem{Han20}
Tengda Han, Weidi Xie, and Andrew Zisserman.
\newblock Self-supervised co-training for video representation learning.
\newblock In {\em Neurips}, 2020.

\bibitem{he2020momentum}
Kaiming He, Haoqi Fan, Yuxin Wu, Saining Xie, and Ross Girshick.
\newblock Momentum contrast for unsupervised visual representation learning.
\newblock In {\em Proceedings of the IEEE/CVF Conference on Computer Vision and
  Pattern Recognition}, pages 9729--9738, 2020.

\bibitem{he2016deep}
Kaiming He, Xiangyu Zhang, Shaoqing Ren, and Jian Sun.
\newblock Deep residual learning for image recognition.
\newblock In {\em Proceedings of the IEEE conference on computer vision and
  pattern recognition}, pages 770--778, 2016.

\bibitem{hu2021contrast}
Kai Hu, Jie Shao, Yuan Liu, Bhiksha Raj, Marios Savvides, and Zhiqiang Shen.
\newblock Contrast and order representations for video self-supervised
  learning.
\newblock In {\em Proceedings of the IEEE/CVF International Conference on
  Computer Vision}, pages 7939--7949, 2021.

\bibitem{huang2019ccnet}
Zilong Huang, Xinggang Wang, Lichao Huang, Chang Huang, Yunchao Wei, and Wenyu
  Liu.
\newblock Ccnet: Criss-cross attention for semantic segmentation.
\newblock In {\em Proceedings of the IEEE/CVF International Conference on
  Computer Vision}, pages 603--612, 2019.

\bibitem{DBLP:journals/corr/IsolaZKA15}
Phillip Isola, Daniel Zoran, Dilip Krishnan, and Edward~H. Adelson.
\newblock Learning visual groups from co-occurrences in space and time.
\newblock {\em CoRR}, abs/1511.06811, 2015.

\bibitem{kay2017kinetics}
Will Kay, Joao Carreira, Karen Simonyan, Brian Zhang, Chloe Hillier, Sudheendra
  Vijayanarasimhan, Fabio Viola, Tim Green, Trevor Back, Paul Natsev, et~al.
\newblock The kinetics human action video dataset.
\newblock {\em arXiv preprint arXiv:1705.06950}, 2017.

\bibitem{komodakis2018unsupervised}
Nikos Komodakis and Spyros Gidaris.
\newblock Unsupervised representation learning by predicting image rotations.
\newblock In {\em International Conference on Learning Representations (ICLR)},
  2018.

\bibitem{krizhevsky2012imagenet}
Alex Krizhevsky, Ilya Sutskever, and Geoffrey~E Hinton.
\newblock Imagenet classification with deep convolutional neural networks.
\newblock {\em Advances in neural information processing systems},
  25:1097--1105, 2012.

\bibitem{Kuehne11}
H. Kuehne, H. Jhuang, E. Garrote, T. Poggio, and T. Serre.
\newblock {HMDB}: a large video database for human motion recognition.
\newblock In {\em Proceedings of the International Conference on Computer
  Vision (ICCV)}, 2011.

\bibitem{liu2022tcgl}
Yang Liu, Keze Wang, Lingbo Liu, Haoyuan Lan, and Liang Lin.
\newblock Tcgl: Temporal contrastive graph for self-supervised video
  representation learning.
\newblock {\em IEEE Transactions on Image Processing}, 31:1978--1993, 2022.

\bibitem{liu2021swin}
Ze Liu, Yutong Lin, Yue Cao, Han Hu, Yixuan Wei, Zheng Zhang, Stephen Lin, and
  Baining Guo.
\newblock Swin transformer: Hierarchical vision transformer using shifted
  windows.
\newblock {\em arXiv preprint arXiv:2103.14030}, 2021.

\bibitem{liu2020teinet}
Zhaoyang Liu, Donghao Luo, Yabiao Wang, Limin Wang, Ying Tai, Chengjie Wang,
  Jilin Li, Feiyue Huang, and Tong Lu.
\newblock Teinet: Towards an efficient architecture for video recognition.
\newblock In {\em Proceedings of the AAAI Conference on Artificial
  Intelligence}, volume~34, pages 11669--11676, 2020.

\bibitem{liu2021video}
Ze Liu, Jia Ning, Yue Cao, Yixuan Wei, Zheng Zhang, Stephen Lin, and Han Hu.
\newblock Video swin transformer.
\newblock {\em arXiv preprint arXiv:2106.13230}, 2021.

\bibitem{loshchilov2016sgdr}
Ilya Loshchilov and Frank Hutter.
\newblock Sgdr: Stochastic gradient descent with warm restarts.
\newblock {\em arXiv preprint arXiv:1608.03983}, 2016.

\bibitem{loshchilov2017decoupled}
Ilya Loshchilov and Frank Hutter.
\newblock Decoupled weight decay regularization.
\newblock {\em arXiv preprint arXiv:1711.05101}, 2017.

\bibitem{Misra-2016-5596}
Ishan Misra, C.~Lawrence Zitnick, and Martial Hebert.
\newblock Shuffle and learn: Unsupervised learning using temporal order
  verification.
\newblock In {\em Proceedings of (ECCV) European Conference on Computer
  Vision}, pages 527 -- 544, October 2016.

\bibitem{neimark2021video}
Daniel Neimark, Omri Bar, Maya Zohar, and Dotan Asselmann.
\newblock Video transformer network.
\newblock {\em arXiv preprint arXiv:2102.00719}, 2021.

\bibitem{noroozi2016unsupervised}
Mehdi Noroozi and Paolo Favaro.
\newblock Unsupervised learning of visual representations by solving jigsaw
  puzzles.
\newblock In {\em European conference on computer vision}, pages 69--84.
  Springer, 2016.

\bibitem{oord2018representation}
Aaron van~den Oord, Yazhe Li, and Oriol Vinyals.
\newblock Representation learning with contrastive predictive coding.
\newblock {\em arXiv preprint arXiv:1807.03748}, 2018.

\bibitem{paszke2019pytorch}
Adam Paszke, Sam Gross, Francisco Massa, Adam Lerer, James Bradbury, Gregory
  Chanan, Trevor Killeen, Zeming Lin, Natalia Gimelshein, Luca Antiga, et~al.
\newblock Pytorch: An imperative style, high-performance deep learning library.
\newblock {\em arXiv preprint arXiv:1912.01703}, 2019.

\bibitem{pathak2016context}
Deepak Pathak, Philipp Krahenbuhl, Jeff Donahue, Trevor Darrell, and Alexei~A
  Efros.
\newblock Context encoders: Feature learning by inpainting.
\newblock In {\em Proceedings of the IEEE conference on computer vision and
  pattern recognition}, pages 2536--2544, 2016.

\bibitem{patrick2021keeping}
Mandela Patrick, Dylan Campbell, Yuki~M Asano, Ishan Misra~Florian Metze,
  Christoph Feichtenhofer, Andrea Vedaldi, Jo Henriques, et~al.
\newblock Keeping your eye on the ball: Trajectory attention in video
  transformers.
\newblock {\em Advances in neural information processing systems}, 2012.

\bibitem{DBLP:journals/corr/abs-2008-03800}
Rui Qian, Tianjian Meng, Boqing Gong, Ming{-}Hsuan Yang, Huisheng Wang,
  Serge~J. Belongie, and Yin Cui.
\newblock Spatiotemporal contrastive video representation learning.
\newblock {\em CoRR}, abs/2008.03800, 2020.

\bibitem{recasens2021broaden}
Adri\`a Recasens, Pauline Luc, Jean-Baptiste Alayrac, Luyu Wang, Florian Strub,
  Corentin Tallec, Mateusz Malinowski, Viorica P\u{a}tr\u{a}ucean, Florent
  Altch\'e, Michal Valko, Jean-Bastien Grill, A\"aron van~den Oord, and Andrew
  Zisserman.
\newblock Broaden your views for self-supervised video learning.
\newblock In {\em Proceedings of the IEEE/CVF International Conference on
  Computer Vision (ICCV)}, pages 1255--1265, October 2021.

\bibitem{ridnik2021imagenet}
Tal Ridnik, Emanuel Ben-Baruch, Asaf Noy, and Lihi Zelnik-Manor.
\newblock Imagenet-21k pretraining for the masses.
\newblock {\em arXiv preprint arXiv:2104.10972}, 2021.

\bibitem{sharir2021image}
Gilad Sharir, Asaf Noy, and Lihi Zelnik-Manor.
\newblock An image is worth 16x16 words, what is a video worth?
\newblock {\em arXiv preprint arXiv:2103.13915}, 2021.

\bibitem{simonyan2014two}
Karen Simonyan and Andrew Zisserman.
\newblock Two-stream convolutional networks for action recognition in videos.
\newblock {\em arXiv preprint arXiv:1406.2199}, 2014.

\bibitem{song2018self}
Jingkuan Song, Hanwang Zhang, Xiangpeng Li, Lianli Gao, Meng Wang, and Richang
  Hong.
\newblock Self-supervised video hashing with hierarchical binary auto-encoder.
\newblock {\em IEEE Transactions on Image Processing}, 27(7):3210--3221, 2018.

\bibitem{UCF101}
Khurram Soomro, Amir~Roshan Zamir, and Mubarak Shah.
\newblock {UCF101}: A dataset of 101 human action classes from videos in the
  wild.
\newblock In {\em CRCV-TR-12-01}, 2012.

\bibitem{pmlr-v37-srivastava15}
Nitish Srivastava, Elman Mansimov, and Ruslan Salakhudinov.
\newblock Unsupervised learning of video representations using lstms.
\newblock In {\em Proceedings of the 32nd International Conference on Machine
  Learning}, volume~37 of {\em Proceedings of Machine Learning Research}, pages
  843--852, Lille, France, 07--09 Jul 2015. PMLR.

\bibitem{tao2020self}
Li Tao, Xueting Wang, and Toshihiko Yamasaki.
\newblock Self-supervised video representation learning using inter-intra
  contrastive framework.
\newblock In {\em Proceedings of the 28th ACM International Conference on
  Multimedia}, pages 2193--2201, 2020.

\bibitem{touvron2020training}
Hugo Touvron, Matthieu Cord, Matthijs Douze, Francisco Massa, Alexandre
  Sablayrolles, and Herv{\'e} J{\'e}gou.
\newblock Training data-efficient image transformers \& distillation through
  attention.
\newblock {\em arXiv preprint arXiv:2012.12877}, 2020.

\bibitem{tran2019video}
Du Tran, Heng Wang, Lorenzo Torresani, and Matt Feiszli.
\newblock Video classification with channel-separated convolutional networks.
\newblock In {\em Proceedings of the IEEE/CVF International Conference on
  Computer Vision}, pages 5552--5561, 2019.

\bibitem{vaswani2017attention}
Ashish Vaswani, Noam Shazeer, Niki Parmar, Jakob Uszkoreit, Llion Jones,
  Aidan~N Gomez, Lukasz Kaiser, and Illia Polosukhin.
\newblock Attention is all you need.
\newblock {\em arXiv preprint arXiv:1706.03762}, 2017.

\bibitem{vincent2008extracting}
Pascal Vincent, Hugo Larochelle, Yoshua Bengio, and Pierre-Antoine Manzagol.
\newblock Extracting and composing robust features with denoising autoencoders.
\newblock In {\em Proceedings of the 25th international conference on Machine
  learning}, pages 1096--1103, 2008.

\bibitem{wang2020video}
Heng Wang, Du Tran, Lorenzo Torresani, and Matt Feiszli.
\newblock Video modeling with correlation networks.
\newblock In {\em Proceedings of the IEEE/CVF Conference on Computer Vision and
  Pattern Recognition}, pages 352--361, 2020.

\bibitem{wang2018learning}
Jue Wang and Anoop Cherian.
\newblock Learning discriminative video representations using adversarial
  perturbations.
\newblock In {\em Proceedings of the European Conference on Computer Vision
  (ECCV)}, pages 685--701, 2018.

\bibitem{wang2019gods}
Jue Wang and Anoop Cherian.
\newblock Gods: Generalized one-class discriminative subspaces for anomaly
  detection.
\newblock In {\em Proceedings of the IEEE/CVF International Conference on
  Computer Vision}, pages 8201--8211, 2019.

\bibitem{wang2020self}
Jiangliu Wang, Jianbo Jiao, and Yun-Hui Liu.
\newblock Self-supervised video representation learning by pace prediction.
\newblock In {\em European conference on computer vision}, pages 504--521.
  Springer, 2020.

\bibitem{wang2018non}
Xiaolong Wang, Ross Girshick, Abhinav Gupta, and Kaiming He.
\newblock Non-local neural networks.
\newblock In {\em Proceedings of the IEEE conference on computer vision and
  pattern recognition}, pages 7794--7803, 2018.

\bibitem{7410677}
Xiaolong Wang and Abhinav Gupta.
\newblock Unsupervised learning of visual representations using videos.
\newblock In {\em 2015 IEEE International Conference on Computer Vision
  (ICCV)}, pages 2794--2802, 2015.

\bibitem{wang2020end}
Yuqing Wang, Zhaoliang Xu, Xinlong Wang, Chunhua Shen, Baoshan Cheng, Hao Shen,
  and Huaxia Xia.
\newblock End-to-end video instance segmentation with transformers.
\newblock {\em arXiv preprint arXiv:2011.14503}, 2020.

\bibitem{xu2019self}
Dejing Xu, Jun Xiao, Zhou Zhao, Jian Shao, Di Xie, and Yueting Zhuang.
\newblock Self-supervised spatiotemporal learning via video clip order
  prediction.
\newblock In {\em Proceedings of the IEEE/CVF Conference on Computer Vision and
  Pattern Recognition}, pages 10334--10343, 2019.

\bibitem{yao2021seco}
Ting Yao, Yiheng Zhang, Zhaofan Qiu, Yingwei Pan, and Tao Mei.
\newblock Seco: Exploring sequence supervision for unsupervised representation
  learning.
\newblock In {\em AAAI}, volume~2, page~7, 2021.

\bibitem{zhang2020dynamic}
Li Zhang, Dan Xu, Anurag Arnab, and Philip~HS Torr.
\newblock Dynamic graph message passing networks.
\newblock In {\em Proceedings of the IEEE/CVF Conference on Computer Vision and
  Pattern Recognition}, pages 3726--3735, 2020.

\end{thebibliography}
}

\newpage
% \mbox{~}
% \clearpage

\appendix
\section{Self-Supervised Learning Frameworks}

In the main paper we presented experimental results obtained by implementing our proposed LSTCL procedure under three popular self-supervised constrastive learning frameworks: MoCo v3~\cite{chen2021empirical}, BYOL~\cite{grill2020bootstrap}, and SimSiam~\cite{chen2020exploring}. Since we used the framework of MoCo v3 to present our method (section 4), here we provide a short description of the other two frameworks---BYOL and SimSiam-- and of how we use them in our method.

\textbf{BYOL.} BYOL is a self-supervised learning framework consisting of a momentum-encoder ($f_{\theta_m}$), an online encoder ($f_\theta$) and a predictor MLP ($g_{\theta_p}$) which is connected to the online encoder. The momentum-encoder is the moving average of the online encoder, which is controlled by a momentum parameter $m$. There is no back-propagation into the momentum-encoder parameters $\theta_m$. The momentum update can be written as:
\begin{equation}
    \theta_m = m\theta_m + (1-m)\theta
\end{equation}
Differently from MoCo, BYOL uses only positive pairs of examples. It minimizes the negative cosine similarity over all positive pairs. In our setting, the set of short clips is processed by the online encoder $f_\theta$ to yield a set of ``query'' examples $Q = \{q^1,q^2,...q^B\}$ where $q^i = f_\theta(x_S^i)\in\mathbb{R}^{D}$. The set of long clips is processed by the momentum encoder $f_{\theta_m}$ to produce ``key'' examples $K = \{k^1,k^2,...k^B\}$. Then, the LSTCL method with BYOL loss minimizes the following objective:
\begin{equation}
    \mathcal{L}_{BYOL} = \sum_i [ 2 - 2\cdot\frac{g_{\theta_p}(q^i)^\top k^i}{||g_{\theta_p}(q^i)||_2\cdot||k^i||_2} ].
\end{equation}
We symmetrize the objective by adding to this loss a dual term obtained by reversing the role of the long and the short clips, i.e., by computing queries from long clips with the online encoder ($q^i = f_\theta(x_L^i)$) and keys from the short clips with the momentum encoder ($k^i = f_{\theta_m}(x_S^i)$).

\textbf{SimSiam.} SimSiam can be viewed as the BYOL method without momentum-encoder and momentum update. Thus, in our setting the queries and the keys are computed from the short view and the long view, respectively, but using the same encoder: $q^i = f_\theta(x_S^i)$, $k^i = f_\theta(x_L^i)$. During training SimSiam applies the stop-gradient operation on the key view. Thus, LSTCL with SimSiam loss mimimizes:
\begin{equation}
    \mathcal{L}_{SimSiam} = \sum_i [ 2 - 2\cdot\frac{g_{\theta_p}(q^i)^\top SG(k^i)}{||g_{\theta_p}(q^i)||_2\cdot||SG(k^i)||_2} ]
\end{equation}
where $SG()$ denotes the stop-gradient operation. Even here we symmetrize the objective by adding a dual term with reversed roles for the long and the short clip.

\section{Analysis of Space-Time Swin Transformer}

We note that our Space-Time Swin Transformer (ST Swin) bears relations with the Video Swin model~\cite{liu2021video} which is concurrently presented in unpublished work. Video Swin differs from our ST Swin in the way it treats the temporal dimension. Video Swin subdivides the video volume into 3D neighborhoods for a self-attention computation that is {\it local in both space and time}. Conversely, our ST Swin inflates 2D spatial Swin blocks~\cite{liu2021swin} into space-time attention tubes that cover the entire temporal extent of the clip. This gives our model the ability to compare patches from {\it all} frames within the same spatial neighborhood. Here we present an empirical comparison between Video Swin (using the code provided by the authors) and our ST Swin, with both models pretrained using our LSTCL and then finetuned on K400. Table~\ref{swincomparison} shows that ST Swin achieves better accuracy. We also list the result reported in the original Video Swin paper for the case when this model is pretrained on ImageNet and then finetuned on K400. This result shows that Video Swin trained with our LSTCL achieves higher accuracy than the same model pretrained on ImageNet using longer clips (32 frames instead of 16). 

\begin{table}[!ht]
\footnotesize
\centering
\begin{tabular}{|l|c|c|c|c|}
\hline
Model       &Clip Size & Additional data & Pretraining & Acc. \\ \hline \hline
ST Swin &$16\ \mathrm{x}\ 224^2$  & - & LSTCL  & 81.5\% \\ \hline
Video Swin &$16\ \mathrm{x}\ 224^2$  & - & LSTCL & 81.0\% \\ \hline \hline
Video Swin  &$32\ \mathrm{x}\ 224^2$  &IN-1K & Superv. & 80.6\% \\
\hline
\end{tabular}
\caption{Comparison between ST Swin and Video Swin on K400.}
\label{swincomparison}
\end{table}

\end{document}